\documentclass{article}

    \PassOptionsToPackage{numbers, compress}{natbib}
 \usepackage[preprint]{neurips_2025}

\usepackage[utf8]{inputenc} 
\usepackage[T1]{fontenc}    
\usepackage{hyperref}       
\usepackage{url}            
\usepackage{booktabs}       
\usepackage{amsfonts}       
\usepackage{nicefrac}       
\usepackage{microtype}      
\usepackage{xcolor}         
\usepackage{subcaption}
\usepackage{graphicx}
\usepackage{amsmath}
\usepackage{tabularx}
\usepackage{ltablex}
\usepackage{makecell}
\usepackage{amssymb}
\usepackage{float}

\title{Foresight-England: Development of a National-Scale Generative AI Model of Electronic Health Records for Medical Event Prediction across the COVID-19 Pandemic}

\author{
    Simon Ellershaw\thanks{These authors contributed equally to this work.} \textsuperscript{1, 2}
    \And
    Christopher Tomlinson*\thanks{Corresponding author: \texttt{christopher.tomlinson@ucl.ac.uk}}
    \textsuperscript{1, 2, 3, 4}
    \And
    Zeljko Kraljevic \textsuperscript{2}
    \And
    Spiros Denaxas \textsuperscript{1, 4, 5, 6, 7}
    \And
    Harry Hemingway \textsuperscript{1, 4, 7}
    \And
    Cathie Sudlow \textsuperscript{8}
    \And
    Angela M. Wood \textsuperscript{6, 9, 10, 11, 12, 13, 14}
    \And
    Anoop D. Shah \textsuperscript{1, 4, 15}
    \And
    Richard Dobson \textsuperscript{1, 2, 4, 7, 16}
    \And
    on behalf of the CVD-COVID-UK/COVID-IMPACT Consortium\\
    \AND 
\begin{minipage}{\textwidth} 
    \vspace{0.5cm} 
\normalfont \textsuperscript{1} Institute of Health Informatics, University College London, London, UK \\
\normalfont \textsuperscript{2} Department of Biostatistics and Health Informatics, Institute of Psychiatry, Psychology and Neuroscience, King’s College London, London, UK \\
\normalfont \textsuperscript{3} King's Institute for Artificial Intelligence, King’s College London, London, UK \\
\normalfont \textsuperscript{4} University College London Hospitals National Institute for Health Research Biomedical Research Centre, London, UK \\
\normalfont \textsuperscript{5}  Interdisciplinary Transformation University, Linz, Austria \\
\normalfont \textsuperscript{6} British Heart Foundation Data Science Centre, Health Data Research UK, London, UK \\
\normalfont \textsuperscript{7} Health Data Research UK, London, UK \\
\normalfont \textsuperscript{8} Usher Institute, School of Population Health Sciences, The University of Edinburgh \\
\normalfont \textsuperscript{9} British Heart Foundation Cardiovascular Epidemiology Unit, Department of Public Health and Primary Care, University of Cambridge, Cambridge, UK \\
\normalfont \textsuperscript{10} Victor Phillip Dahdaleh Heart and Lung Research Institute, University of Cambridge, Cambridge, UK \\
\normalfont \textsuperscript{11} British Heart Foundation Centre of Research Excellence, University of Cambridge, Cambridge, UK \\
\normalfont \textsuperscript{12} National Institute for Health and Care Research Blood and Transplant Research Unit in Donor Health and Behaviour, University of Cambridge, Cambridge, UK \\
\normalfont \textsuperscript{13} Health Data Research UK Cambridge, Wellcome Genome Campus and University of Cambridge, Cambridge, UK \\
\normalfont \textsuperscript{14} Cambridge Centre of Artificial Intelligence in Medicine, University of Cambridge, Cambridge, UK \\
\normalfont \textsuperscript{15} Department of Clinical Pharmacology, University College London Hospitals NHS Foundation Trust, London, UK \\
\normalfont \textsuperscript{16} National Institute for Health Research Biomedical Research Centre at South London and Maudsley NHS Foundation Trust and King's College London, London, UK \\
\end{minipage}
}

\begin{document}

\maketitle

\newpage
\begin{abstract}

Foresight-England (Foresight-E) is the first national-scale generative foundation model of electronic health records (EHRs), developed as a research pilot strictly for COVID-19-related research. We evaluated its ability to model the direct and indirect effects of the COVID-19 pandemic. 
\newline
\newline
Trained from scratch entirely within the NHS England Secure Data Environment, Foresight-E is a 243-million-parameter transformer decoder. It is trained and evaluated on a national-scale, de-identified, longitudinal EHR dataset of approximately 61 million individuals, integrating primary and secondary care, death registrations, and COVID-19 testing/vaccination datasets. Training and validation were conducted on a random subset of 90\% of individuals (54.9 million) for events recorded between 1st November 2018 and 31st December 2022. The remaining 10\% of individuals (6.1 million) were held out for evaluation.
\newline
\newline
Foresight-E models patient timelines autoregressively, predicting the next medical event given an individual's prior history. At inference, it operates zero-shot, generating predictions for any concept in its approximately 40,000-code medical vocabulary without additional task-specific training. Our custom tokenisation scheme retains the recorded clinical granularity of ICD-10, OPCS-4, and SNOMED CT codes, and jointly represents both absolute (e.g., calendar dates) and relative timing (e.g., chronological age). 
\newline
\newline
We designed and implemented an evaluation framework spanning 30-day COVID-19 hospitalisation and mortality using Brier scores and area under the receiver operating characteristic (AUROC) and precision-recall (AUPRC) curves. Subgroup analysis of these results by age, ethnicity, sex and COVID vaccination status was also conducted. We also tested Foresight-E on medical events from 2023, extending beyond its 2018–2022 training period, to assess how well it captured the enduring, system-wide indirect effects of the pandemic on future unseen data, simulating a prospective deployment. We benchmarked model performance against logistic regression and XGBoost. 
\newline
\newline
As detailed in the Project Status section, NHS England has paused access to data for the Foresight-E project, meaning quantitative results are not currently available. Instead, we share our strategy for tokenisation, model architecture, training, inference, and evaluation, as a methodological template and a case study in the challenges of building population-scale EHR foundation models.
\end{abstract}

\newpage

\section{Lay summary}
Predicting who gets ill, when, and with which diseases are vital questions for individuals, doctors, and healthcare systems, such as the National Health Service (NHS). The COVID-19 pandemic urgently highlighted this need. The NHS had to quickly identify people who might become very sick if they caught the virus, so they could be prioritised for vaccinations or treatments. While many predictive tools, including some using artificial intelligence (AI), were developed during the pandemic, they often focused on single, narrow questions, like predicting the risk of death only \textit{after} a patient was admitted to hospital. Because they couldn't look at the bigger picture, they struggled to capture the wider, long-term impacts of the pandemic on the healthcare system.

To address this, we developed a new AI system called Foresight-England (Foresight-E) to predict a patient’s future medical events. Our tool learns from past patient data to estimate what might happen to a patient's health in the future, and when these events might occur. It works a bit like the predictive text on a mobile phone, which tries to predict the next word in a sentence, but it uses medical codes instead of words.

Our goal was to test Foresight-E's ability to model changes in people's health during the pandemic. We looked at two main areas:

\begin{itemize}
    \item \textbf{Direct Effects of COVID-19:} We tested how accurately the AI could tell us which patients would be hospitalised or die within 30 days of a positive COVID-19 test. We also checked if the tool could learn and adapt to the shifting nature of the pandemic, such as new viral variants and the vaccine rollout. Importantly, we checked if the AI worked fairly across different demographic groups, looking for evidence of bias in the underlying data that might lead to unequal predictions.
    \item \textbf{Indirect Effects of the Pandemic:} The pandemic caused severe disruptions to routine healthcare, leaving lasting effects. To see if the tool could predict these broader impacts, we trained it on data up to the end of 2022, and then tested it on "unseen" data from 2023. We looked at emergency hospital admissions, overall deaths, and the new onset of over 1,400 different diseases whose diagnosis or treatment might have been delayed or triggered by the pandemic.
\end{itemize}

AI models are only as good as the data they learn from. To make sure the model was fair and represented people from all backgrounds, we trained Foresight-E using de-identified electronic health records from 54.9 million people—representing almost the entire population of England. We then tested the tool on a separate group of 6.1 million people. This data was securely accessed via the British Heart Foundation Data Science Centre’s CVD-COVID-UK/COVID-IMPACT consortium. The data looked a bit like a computer spreadsheet of medical codes and dates, not written text notes from doctors. It included GP records, hospital visits, COVID-19 testing, vaccinations, and national death registrations.

The project was designed so that the data, the AI model, and its predictions were kept entirely within a highly secure digital system called the NHS England Secure Data Environment (SDE). This environment uses strict rules (the "Five Safes" framework) meaning no patient data ever left the NHS. The AI was built from scratch inside this secure system. Our industry partners, Amazon Web Services (AWS) and Databricks, provided computer power and technical help, but they had absolutely no access to the data, the AI model, or control over the research. It is important to note that Foresight-E was developed strictly as a research project for COVID-19 and is not used by the NHS for patient care.

NHS England has paused access to the data for this project. Because we cannot access the secure environment, we cannot retrieve the initial results of our study. Therefore, we have provided a transparent overview of how we designed and tested the system, alongside "placeholder" mock-ups of our results tables to demonstrate the work we have already completed, and how we were intending to report it.

\newpage

\section{Project Status}
\label{section_status}

In May 2025, the British Medical Association and Royal College of General Practitioners’ Joint GP IT Committee (JGPITC) raised concerns that they were unaware that primary care data collected for COVID-19 research was being used to train an AI model, and queried whether the correct processes, including GDPR principles, had been followed \cite{bmj_foresight}. Following these concerns, NHS England paused access to data for the Foresight-E project whilst a governance review was carried out. This review considered whether the correct approval processes and privacy protections were in place for a project of this nature. The JGPITC additionally wrote to the Information Commissioner's Office (ICO), the UK’s data protection regulator, asking them to investigate. 

In April 2026, following a thorough review, the ICO closed its review of the project and concluded that the project's data use was compatible with the purpose for which it was originally shared therefore there was no breach of GDPR. The project's access to the NHS England SDE remains paused.

At the point at which data access was paused, the researchers had trained several iterations of the Foresight-E model (including on the full training dataset), generated multiple sets of predictions for direct and indirect COVID-19 outcomes in the test set, and run initial quantitative evaluations on these predictions, producing aggregated performance metrics, as detailed in this manuscript. However these aggregate predictions had not yet been requested for export from the NHS England Secure Data Environment (SDE) via the approved Safe Output Service, subject to statistical disclosure control and review. 

This means quantitative results are not available, and this paper instead seeks to provide a transparent overview of the underpinning methodology, evaluation strategy and work undertaken to date on Foresight-E. Placeholder results are presented to provide full transparency over the nature of the evaluation and aggregate data (e.g. tables, figures) that were intended to be exported from the SDE following standard procedure. Whilst the researchers describe the work to the best of their ability, it is important to note that since data access was paused, they have been unable to access not only the underlying data, but their codebase, models, experiment tracking, documentation and results, which are stored inside the SDE in platforms such as Databricks, Gitlab and MLflow.

\newpage

\section{Introduction}
\label{section_intro}
Accurate risk stratification is an important component of clinical decision-making, enabling individuals, healthcare professionals and health systems to anticipate adverse outcomes, tailor interventions, and inform resource allocation. Traditional clinical risk prediction models typically combine established expert knowledge with rule-based or statistical approaches that harness a limited number of static features, such as demographics, pre-existing conditions, or test results, at a single point in time to predict a single outcome \cite{qrisk, chad_vasc}. While effective for narrow, well-understood diseases, these methods face fundamental limitations during a global pandemic caused by a novel pathogen. When COVID-19 emerged, healthcare systems lacked a priori knowledge of interacting risk factors, and the epidemiology shifted rapidly with viral variants and changing public health interventions, such as vaccination rollouts. Traditional approaches struggle to model this complexity and cannot easily scale to capture the multitude of ways a systemic shock like COVID-19 impacts the healthcare system across a vast array of possible clinical outcomes. 

Neural network transformer models, initially developed for natural language processing (NLP) \cite{attention_is_all_you_need}, offer a way to harness the temporal and longitudinal information in electronic health records (EHRs), often underutilised by traditional modelling approaches. The move toward generative pretrained transformers (GPTs) using autoregressive next-token prediction has resulted in large language models (LLMs) capable of zero- and few-shot performance across diverse tasks without task-specific retraining \cite{gpt3}, inspiring analogous efforts in healthcare \cite{behrt, medbert, foresight, ethos, delphi, epic_ehr_model}. 

Generative EHR models hold the promise of learning directly from rich, longitudinal data without relying on predefined rules, adapting to shifting epidemiology and capturing complex temporal associations that prove challenging for traditional models - properties which make them uniquely suited to modelling the complexity of COVID-19 pandemic. In applied use they offer the potential to support early detection, risk stratification, and simulation of clinical scenarios in a single model, without the need to retrain for each outcome of interest, enabling faster insights in a rapidly evolving public health emergency.

In England, the Control of Patient Information (COPI) Regulations \cite{copi} provided a legal framework to make de-identified, routinely collected, national-scale NHS data available for COVID-19-related research \cite{wood2021linked}. These data include primary care, secondary care, COVID-19 testing, vaccination, and mortality data from the population of England (approximately 61 million people \cite{pineda2024ethnicity}) and are securely stored within the NHS England Secure Data Environment (NHSE SDE) \cite{nhs_sde}. Despite the potential of such a resource for the development and evaluation of AI models for COVID-19 research, computational restrictions within the NHSE SDE have limited previous projects to substantially smaller cohorts \cite{handy2021,allery2023}. 

In this work, we developed Foresight-England (Foresight-E), a 243-million-parameter transformer trained from scratch entirely within the NHSE SDE on linked primary and secondary care data from 54.9 million people (90\%; total dataset size of 61 million). Foresight-E was designed for the zero-shot prediction of both the direct effects of COVID-19 (e.g., hospitalisation, mortality) and its indirect systemic impacts across the population.

Access to this scale and demographic diversity is important to accurately model the COVID-19 pandemic, particularly to capture the outcomes of ethnic minority groups and patients with rare diseases or COVID-related complications who are only represented in statistically significant numbers at a population level \cite{pineda2024ethnicity, thygesen2025rare, knight2022association}. Recent methodological research also shows that EHR foundation models exhibit scaling laws similar to those seen with LLMs, suggesting that larger scale training data improves performance \cite{ehr_scaling}. Linking primary care records with secondary care, COVID-19 testing, vaccination data, and death registries provides a mechanism to capture the entire spectrum of the disease—from mild, community-managed presentations to critical hospital care. Crucially, this national-scale baseline allows for rigorous evaluation of algorithmic fairness across different ethnicities, age groups, and socioeconomic backgrounds.

Indeed, most prior generative EHR models have been restricted to single healthcare institutions \cite{ethos}, specific EHR providers \cite{epic_ehr_model}, or limited population subsets \cite{behrt}. This limits their applicability to the English general population and risks algorithmic bias, a failure to maintain consistent performance across diverse demographic and clinical groups \cite{STANDING}. 

Furthermore, the true burden of COVID-19 extends far beyond acute viral infection. The pandemic caused profound, systemic disruptions to routine healthcare delivery, resulting in delayed diagnoses, altered treatment pathways, and the emergence of post-acute sequelae (such as Long COVID) which continue to impact both individuals and healthcare systems today. Whilst transformer models offer the potential to capture these changes during their training, the extent to which this generalises to future, unseen data remains unquantified. Therefore to rigorously evaluate a model's capacity to capture these critical indirect pandemic effects, and its ability to generalise to future data, it is necessary to test its predictive performance across the wider spectrum of human disease and beyond its training data, here encompassing over 1,400 clinical phenotypes in the year of 2023.

As outlined in Project Status \ref{section_status}, quantitative results are unavailable, therefore we report the data pipeline, tokenisation, architecture, training, inference, and evaluation framework underpinning the model, assessed against the TRIPOD+AI \cite{tripodAI} and PROBAST-AI \cite{probastAI} reporting guidelines (see Appendix \ref{section_tripod} and \ref{section_PROBAST}).

We contribute the following:
\begin{enumerate}
    \item Development of Foresight-E, the first national-scale generative foundation model of EHRs for COVID-19 research.
    \item A reproducible methodology for longitudinal EHR tokenisation and model development.
    \item A comprehensive evaluation framework to assess zero-shot prediction of COVID-19’s direct and indirect effects on a held-out disjoint test set. Comprising 6.1 million unique patients whose data was not used during model training, and temporally held out data.
    \item First demonstration of a multi graphics processing unit (GPU)-accelerated foundation model pretrained on national-scale NHS data within the NHSE SDE.
\end{enumerate}

Together, these provide both a methodological blueprint for future EHR foundation models and a case study in the challenges of developing national-scale generative AI within secure health data environments.

\newpage
\section{Methods}

We developed Foresight-E for COVID-19 research using linked, de-identified, routinely collected national datasets \cite{thygesen2022covid, wood2021linked}. A key principle of Foresight-E is that the model is treated with the same security as the underlying data. Therefore, all processing, model training, inference, and evaluation occurred entirely within the ‘Five Safes’ framework of the NHSE SDE \cite{nhs_sde}. Neither the model weights nor any generated patient timelines can be exported outside this environment; only aggregated, non-disclosive evaluation metrics are eligible for release, via the approved Safe Output Service, subject to statistical disclosure control and review. 

Here we outline the datasets, patient timeline construction, tokenisation, model architecture, training regime, and our inference and evaluation strategy, including uncertainty estimation and comparative baselines. Each step was designed to support robust, generalisable, zero-shot medical event prediction for the direct and indirect effects of COVID-19. 

\subsection{Data}
\label{section_methods_data}
\begin{figure}[h!]
    \centering
    \begin{subfigure}{0.48\textwidth}
        \centering
        \includegraphics[width=\linewidth]{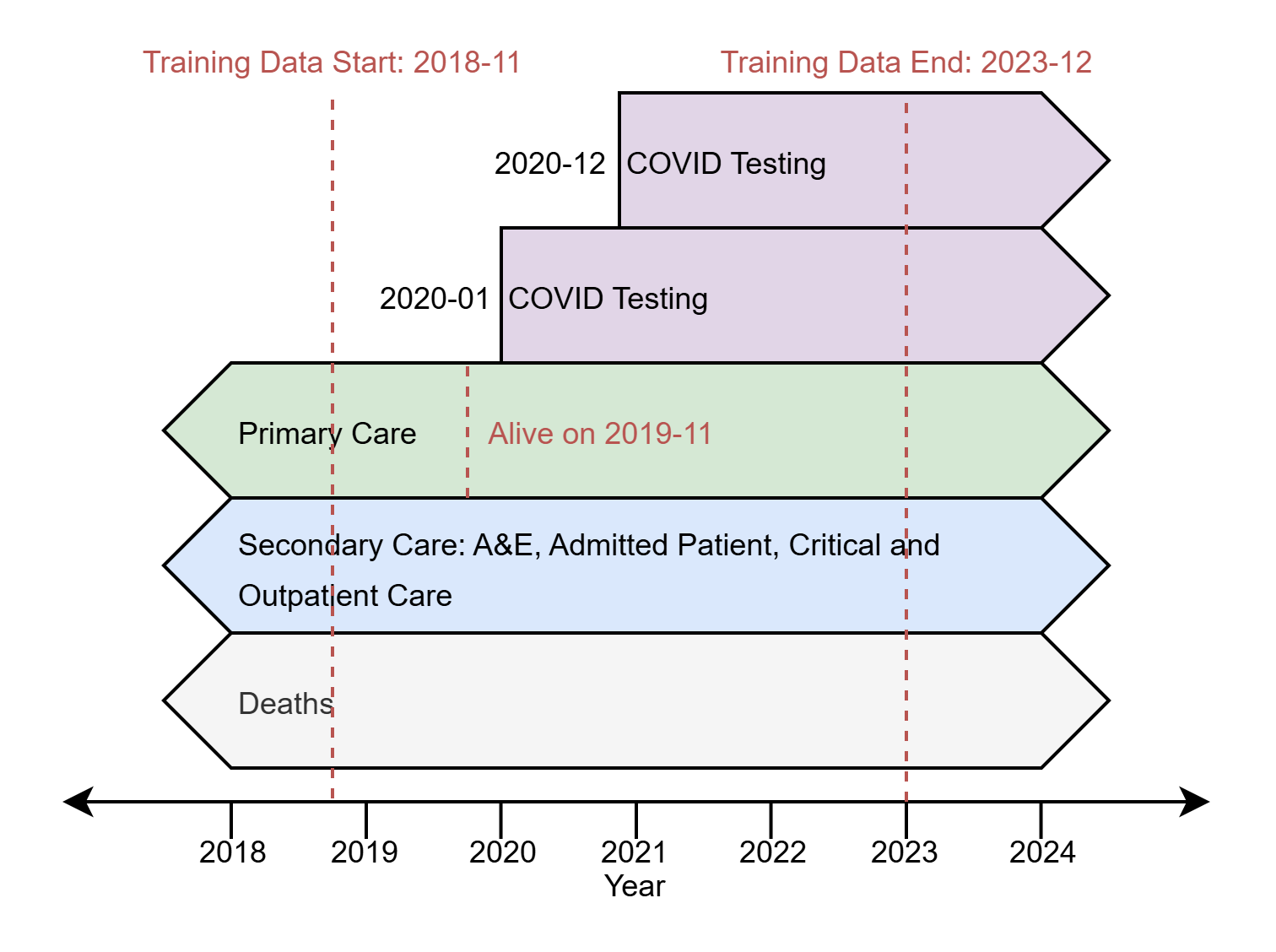}
        \caption{Linked primary/secondary care, deaths, testing, and vaccination datasets (1~Nov~2018–31~Dec~2023).}
        \label{fig_dataset}
    \end{subfigure}
    \hfill
    \begin{subfigure}{0.48\textwidth}
        \centering
        \includegraphics[width=\linewidth]{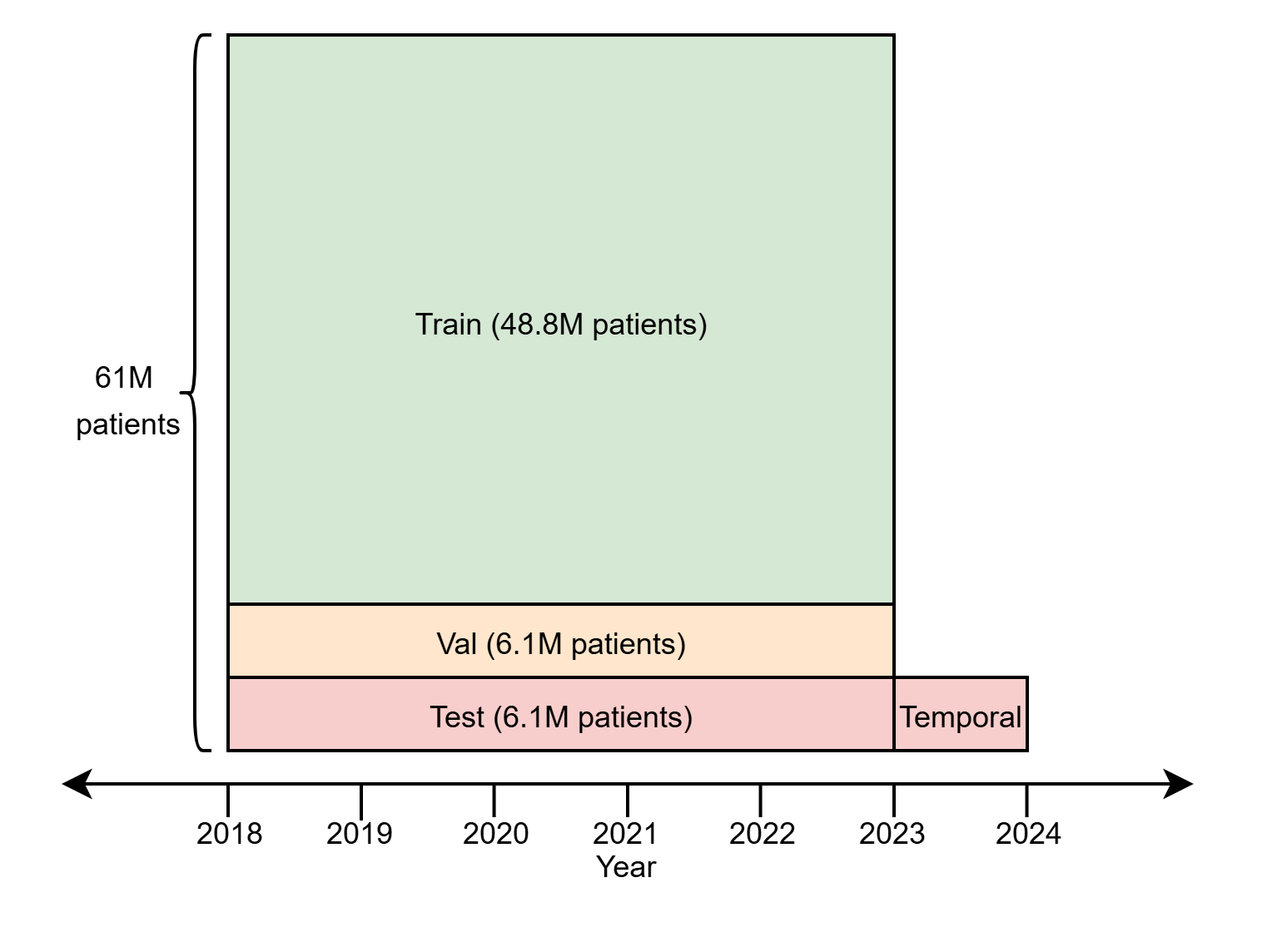}
        \caption{Disjoint 10\% validation/test cohorts; 2023 held out for temporal generalisation evaluation. }
        \label{fig_data_split}
    \end{subfigure}
    \caption{Datasets and cohort splits used in the training and evaluation of Foresight-E. Total sample sizes for each split are derived from previously reported counts~\cite{pineda2024ethnicity}.}
    \label{fig:datasets_combined}
\end{figure}

To train and evaluate Foresight-E, longitudinal patient data were accessed from eight pseudonymised, linked, routinely collected national datasets covering 1~November~2018 to 31~December~2023 \cite{thygesen2022covid, wood2021linked}. These encompassed primary care (General Practice Extraction Service (GPES) Data for Pandemic Planning and Research (GDPPR) \cite{primary_care_data}), secondary care (Hospital Episode Statistics (HES): Outpatients (OP), Accident \& Emergency (A\&E), Admitted Patient Care (APC) and Critical Care (CC) \cite{secondary_care_data}), mortality (Office for National Statistics (ONS) Civil Registration of Deaths \cite{deaths_data}), COVID-19 testing (UK Health Security Agency (UKHSA), formerly Public Health England (PHE), COVID-19 Second Generation Surveillance System (SGSS) \cite{covid_data_1}) and COVID-19 vaccination (NHS England COVID-19 Vaccination Status \cite{covid_data_2}) data (see Fig.~\ref{fig_dataset}).

The base cohort comprised individuals alive on or after 1~November~2019 (the GDPPR dataset inclusion criterion) with known age and sex, resident in England, GP-registered, and without conflicting death dates. To ensure a minimum one year of past medical history before predicting outcomes, we included data from 1~November~2018 onwards. Because primary care registration dates were unavailable to confirm exact prior follow-up lengths, this fixed calendar gap served as a population-level proxy for baseline history. To accommodate a dynamic cohort, individuals born after the study start date could enter the cohort at birth, however to enforce the minimum one year history for these newborns, we required them to reach at least one year of age prior to death or the end of the training period (31~December~2022). Ultimately, this yielded a nationally representative cohort of 61 million patients \cite{pineda2024ethnicity}.

\begin{table}[h!]
\caption{\textbf{PLACEHOLDER} Selected characteristics of the training, validation, and test sets (used in Sections \ref{subsection_results_covid} and \ref{subsection_results_1_year}). Ages are computed at the time of the last input event in each patient’s timeline. Counts are rounded to the nearest five to comply with the NHSE SDE's disclosure control requirements. Total sample sizes for each split are derived from previously reported counts~\cite{pineda2024ethnicity}. The rationale for placeholders is provided in Section~\ref{sect:results}.}
\label{table_dataset}
\centering
\begin{tabular}{lcccc}
\hline
\toprule
\textbf{} & \textbf{Train} & \textbf{Validation} & \textbf{Test: 30-day COVID-19} & \textbf{Test: 1 Year} \\
\midrule
\textbf{Sex} & & & & \\
Female & $\_$ ($\_$\%) & $\_$ ($\_$\%) & $\_$ ($\_$\%) & $\_$ ($\_$\%) \\
Male & $\_$ ($\_$\%) & $\_$ ($\_$\%) & $\_$ ($\_$\%) & $\_$ ($\_$\%) \\
\midrule
\textbf{Ethnicity} & & & & \\
Asian & $\_$ ($\_$\%) & $\_$ ($\_$\%) & $\_$ ($\_$\%) & $\_$ ($\_$\%) \\
Black & $\_$ ($\_$\%) & $\_$ ($\_$\%) & $\_$ ($\_$\%) & $\_$ ($\_$\%) \\
Mixed & $\_$ ($\_$\%) & $\_$ ($\_$\%) & $\_$ ($\_$\%) & $\_$ ($\_$\%) \\
Other& $\_$ ($\_$\%) & $\_$ ($\_$\%) & $\_$ ($\_$\%) & $\_$ ($\_$\%) \\
Unknown & $\_$ ($\_$\%) & $\_$ ($\_$\%) & $\_$ ($\_$\%) & $\_$ ($\_$\%) \\
White & $\_$ ($\_$\%) & $\_$ ($\_$\%) & $\_$ ($\_$\%) & $\_$ ($\_$\%) \\
\midrule
\textbf{Age} & & & & \\
0-9 & $\_$ ($\_$\%) & $\_$ ($\_$\%) & $\_$ ($\_$\%) & $\_$ ($\_$\%) \\
10-19 & $\_$ ($\_$\%) & $\_$ ($\_$\%) & $\_$ ($\_$\%) & $\_$ ($\_$\%) \\
20-29 & $\_$ ($\_$\%) & $\_$ ($\_$\%) & $\_$ ($\_$\%) & $\_$ ($\_$\%) \\
30-39 & $\_$ ($\_$\%) & $\_$ ($\_$\%) & $\_$ ($\_$\%) & $\_$ ($\_$\%) \\
40-49 & $\_$ ($\_$\%) & $\_$ ($\_$\%) & $\_$ ($\_$\%) & $\_$ ($\_$\%) \\
50-59 & $\_$ ($\_$\%) & $\_$ ($\_$\%) & $\_$ ($\_$\%) & $\_$ ($\_$\%) \\
60-69 & $\_$ ($\_$\%) & $\_$ ($\_$\%) & $\_$ ($\_$\%) & $\_$ ($\_$\%) \\
70-79 & $\_$ ($\_$\%) & $\_$ ($\_$\%) & $\_$ ($\_$\%) & $\_$ ($\_$\%) \\
80-89 & $\_$ ($\_$\%) & $\_$ ($\_$\%) & $\_$ ($\_$\%) & $\_$ ($\_$\%) \\
90+ & $\_$ ($\_$\%) & $\_$ ($\_$\%) & $\_$ ($\_$\%) & $\_$ ($\_$\%) \\
\midrule
\textbf{Total} & 48.8M & 6.1M & 6.1M & $\_$ \\
\bottomrule
\end{tabular}
\end{table}

\newpage

We created training (48.8M patients), validation (6.1M patients), and test sets (6.1M patients) via disjoint 10\% patient samples. All 2023 events were reserved to test temporal generalisation, mimicking a prospective deployment on future, unseen data. Fig.~\ref{fig:datasets_combined} shows temporal coverage and partitioning of the datasets.

\subsection{Patient Timelines}
\label{section_methods_timelines}

Each patient’s history was encoded as a chronological sequence of dated, coded events, including diagnoses, procedures, medications, and other healthcare interactions. Clinical codes followed standard terminologies: ICD-10 \cite{icd} for HES diagnoses and ONS causes of death, OPCS-4 \cite{nhs_clinical_classification} for HES procedures, and SNOMED CT \cite{snomed_ct} for GDPPR records and COVID-19 vaccinations. We supplemented these with custom tokens for events not coded in standard terminologies, such as positive SARS-CoV-2 tests, hospital admission indicators, and primary or secondary diagnosis flags. No free text terms were available in the underlying data, or included in the model training.

Because event timestamps were available only at day-level precision, we applied a consistent within-day ordering: COVID-19 vaccination; SARS-CoV-2 test; GDPPR events; HES OP; HES A\&E; HES APC; HES CC; and, lastly, death. Where relevant, within each dataset events were further ordered by admission date, then by primary diagnoses, secondary diagnoses, procedures, and finally alphanumerically.

Geographic (region) and deprivation (Indices of Multiple Deprivation) data were excluded from the training data to avoid explicitly encoding these existing biases and enable held-out subgroup analysis \cite{STANDING}. Codes deemed sensitive (e.g. relating to sexually transmitted infections) were removed, using a clinically-informed codelist supplied by NHS England \cite{gpdr_data_provision_notice}.

\subsection{Tokenisation}
\label{section_tokenisation}

We converted each patient's event timeline into a sequence of discrete tokens representing clinical codes and temporal intervals. Each sequence began with static demographic tokens for sex (e.g., \texttt{SEX\_FEMALE}) and ethnicity (e.g., \texttt{ETHNICITY\_ASIAN}). We then added the sequence of clinical codes, ordered as described in Section~\ref{section_methods_timelines}, from each patient's timeline as tokens. If two consecutive events occurred on different days, we inserted a time-difference token (e.g., \texttt{TIME\_DIFFERENCE\_1}) to represent the interval in days; no time-difference token was added between consecutive same-day events. To encode absolute calendar time, we inserted a \texttt{YEAR\_START} token at the beginning of each calendar year, followed by an \texttt{AGE\_N} token for the patient’s integer age (or \texttt{AGE\_UNBORN} if not yet born). These yearly markers ensured that time gaps never exceeded 366 days and allowed the representation of age-related and absolute temporal patterns to generalise temporally. 

A lookup-based tokeniser (acting as a simple 1:1 dictionary) mapped all unique tokens in the training data to integer IDs, including clinical codes, age (integer years and unborn), 1–366 day gaps, and special tokens for padding (\texttt{<PAD>}) and sequence end (\texttt{<EOS>}). The resulting fixed tokeniser vocabulary contained approximately 40,000 tokens. At inference time, if a code was encountered that was not in the trained tokeniser’s vocabulary, it was dropped from the sequence without substitution, rather than using an unknown token (e.g., \texttt{<UNK>}).

During training, we truncated each individual patient’s timeline to a maximum of 1,024 tokens using left truncation (discarding the earliest events) to retain the most recent clinical context. Truncation was applied at the event level followed by re-tokenisation, rather than truncating token sequences directly; this preserved demographic, year-start and age tokens and correctly recomputed time-difference tokens. For batching, we right-padded sequences to a uniform length with padding tokens, which were masked from attention and loss calculations. At inference, we pre-truncated inputs to $1{,}024 - L_{\text{forecast}}$, where $L_{\text{forecast}}$ denotes the tokens allocated for forecast generation, instead of implementing dynamic on-GPU truncation. The maximum sequence length of 1,024 tokens was selected to balance capturing the maximum possible clinical context while maintaining adequate training and inference batch sizes on the available NVIDIA A10 GPUs.

\subsection{Training}
\label{section_methods_training}

Foresight-E is a 243-million-parameter transformer-decoder model trained to predict the next code in a tokenised EHR sequence, see Section \ref{section_training_obj}. This is analogous to how large language models (LLMs) are trained and can be considered as predicting what will happen next to a patient, based on their past medical history \cite{gpt3,foresight,ethos}. We adapted the open source Llama 2 architecture \cite{llama2,llama2_hf} and trained the model from scratch with randomly initialised weights due to the inability to import pretrained models into the NHS SDE and the use of a custom vocabulary, with clinical codes and custom tokens, rather than natural language. Training comprised a single pass through the dataset with periodic checkpointing; the checkpoint with the lowest validation loss was selected as the final model. Hyperparameters followed established defaults \cite{foresight,nanoGPT}. Complete architectural and training details are provided in Section \ref{section_methods_training_append}.

\begin{figure}[h!]
\centering
\includegraphics[width=\linewidth]{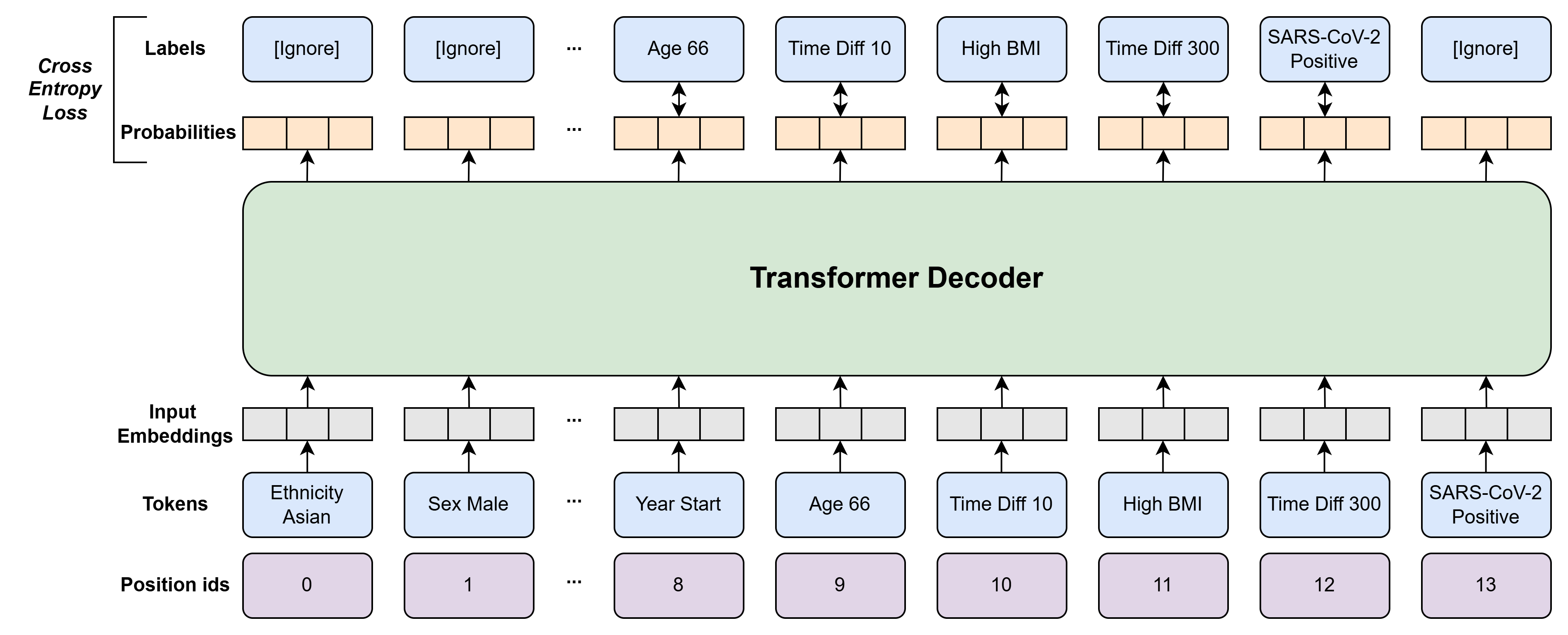}
\caption{Schematic of next token prediction training of Foresight-E on a synthetic patient timeline. Tokens at position IDs 2–7 are omitted for clarity. Given the sequence of tokens up to position $i$, the model predicts the probability distribution over the vocabulary for the token at position $i+1$. These probabilities are compared with the true next token to compute the cross-entropy loss for self-supervised training. Note that the initial static demographic tokens are masked ([Ignore]) during the loss calculation to prevent the model from learning to predict sequence progression solely from demographic features. The final and padding tokens are also ignored because they have no label to compare against.}
\label{fig_model_training}
\end{figure}

\subsection{Zero-Shot Inference}
\label{section_inference}

As a generative model, once trained, Foresight-E can predict future clinical events across the breadth of its vocabulary from patient histories, without requiring task-specific supervised fine-tuning (`zero-shot’ inference). This can be viewed as analogous to LLMs `autocompleting' a sentence. 

\begin{figure}[h!]
\centering
\includegraphics[width=\linewidth]{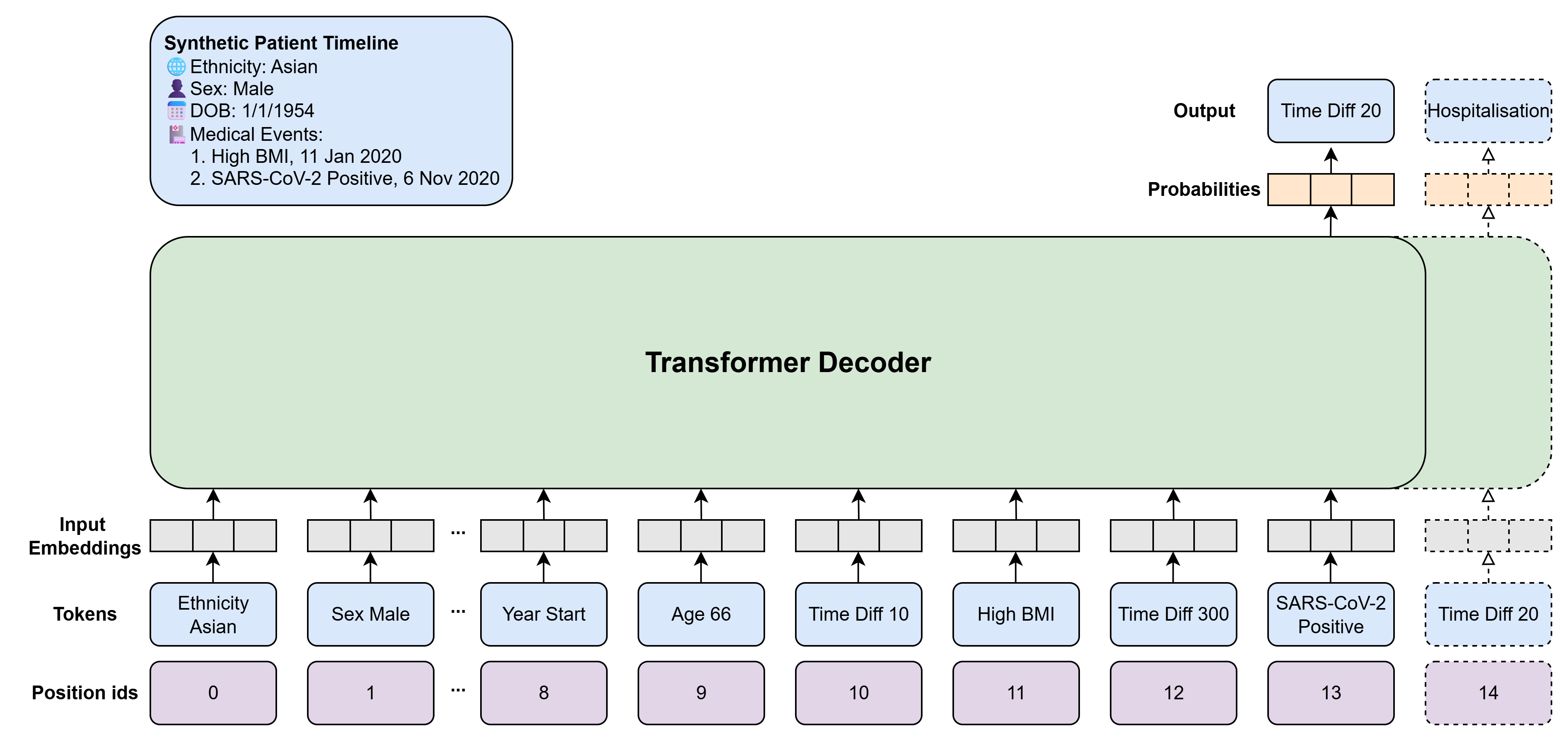}
\caption{Schematic of idealised zero-shot inference on a synthetic patient timeline. Given a history ending with a positive SARS-CoV-2 test, Foresight-E predicts a hospital admission within the next 30 days. Tokens 2–7 are omitted for clarity.}
\label{fig_model_inference}
\end{figure}

At inference, the model is prompted with a tokenised patient timeline (see Section~\ref{section_tokenisation}) and produces a probability distribution over the next token. We sample one token from this distribution and append it to the input. This autoregressive procedure is repeated until a prespecified stopping criterion is met: a sentinel event token (e.g., mortality), a maximum forecast horizon (e.g., 30 days or 1 year), or a limit on newly generated tokens (e.g., 300). The generated tokens are then decoded to form the predicted event timeline, see Figure~\ref{fig_model_inference}

Because greedy decoding yielded repetitive sequences, we used multinomial sampling from the model's output distribution to generate diverse and clinically plausible trajectories. We did not use parameter-dependent sampling techniques (e.g., top-k, temperature), which would have introduced additional task-specific hyperparameters. A key-value (KV) cache was used to increase generation throughput by avoiding the recomputation of past steps.

To quantify predictive uncertainty at the individual patient level, an essential consideration for safe and reliable clinical decision-making~\cite{riley2025uncertainty}, we performed $S=48$ independent stochastic rollouts per patient. This was the maximum batch size possible on a single A10 GPU~\cite{nvidia_a10} given Foresight-E's size and context length. Inference was parallelised across eight GPUs for throughput~\cite{g5_cluster}.

For event $i$ and horizon $t$ (in days), the predicted probability was estimated as~\cite{ethos}

\begin{equation}
P(\text{Event}_{i,t}) = \frac{s_{i,t}}{S},
\end{equation}

where $s_{i,t}$ denotes the number of rollouts in which the event occurred within the horizon. This formulation provides probability estimates across all medical events observed during training, enabling probabilistic zero-shot predictions across diverse clinical outcomes.

\begin{figure}[h!]
\centering
\includegraphics[width=\linewidth]{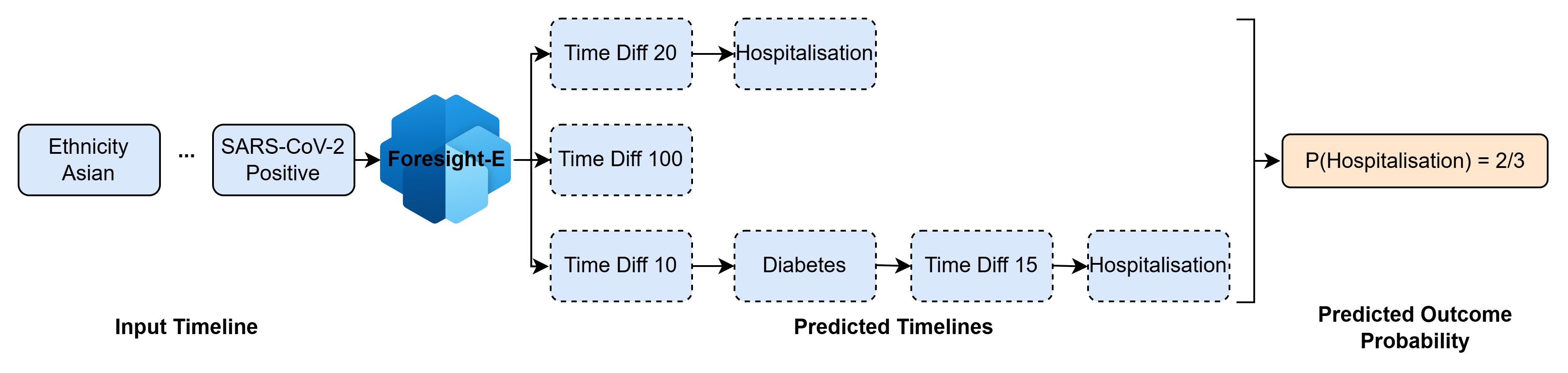}
\caption{Visualisation of an idealised Foresight-E zero-shot inference for the task of 30-day hospitalisation following a positive SARS-CoV-2 test. A tokenised input timeline, shortened for visualisation, is used to prompt three independent predicted timelines in this example. `Time Diff` tokens are in units of days. In two of the predicted timelines, hospitalisation is predicted within the next 30 days (at 20 and 25 days, respectively). The remaining timeline predicts the next event at 100 days, falling outside the target window. Therefore, the predicted probability is 2/3.}
\label{fig_gen_rollout}
\end{figure}

\subsection{Evaluation}
\label{sect:method_outcomes}

We evaluated Foresight-E on two acute COVID outcomes: hospitalisation and death within 30 days following a positive SARS-CoV-2 test. Eligible patients in the test set had at least one positive test between 23~January~2020, the date of the first recorded COVID-19 case in the UK, and 1~December~2023 \cite{lillie2020novel}. While mass community testing in England ended on 1~April~2022, we deliberately included tests beyond this date \cite{cabinetoffice2022living}. Although testing post-April 2022 is inherently selective, capturing predominantly high-risk, hospitalised, or healthcare worker populations, evaluating our model across this policy shift assesses its robustness to changing clinical testing practices and population risk profiles across the pandemic. Timelines were truncated after the first positive test, and predictions generated for the next 30 days. Hospitalisation was defined as an admission in HES APC (\texttt{APC} token), and death as an entry in the Office for National Statistics (ONS) death registry (\texttt{DEATH} token). Inference was completed in 2 days of wall-clock compute time.

Secondly, we assessed the `indirect' effects of COVID-19 in the held-out year of 2023, evaluating emergency hospitalisation, all-cause mortality, and the onset of over 1,400 Phecode-defined diseases. We evaluated these outcomes across the entire held-out cohort, irrespective of a formally recorded positive SARS-CoV-2 test or COVID-19 diagnosis. This population-wide evaluation accounts for the high levels of prior SARS-CoV-2 infection by 2023, and captures the systemic healthcare disruptions that affected patients across all care pathways \cite{indirect_covid_1, indirect_covid_2, indirect_covid_3, coronavirus_infection_survey}.

Here, timelines were truncated after the 2023 year-start token and subsequent events predicted over a one-year horizon. Emergency admissions were identified via admission method codes in HES APC, and mortality as above. Phenome-wide disease onset was defined by the first occurrence of each Phecode, a previously validated collection of over 1,400 groups of ICD-10 codes (example shown in Table~\ref{table_eval_endpoints}) \cite{phecodes}, recorded in HES APC. Inference was completed in 15 days of wall-clock compute time.

\subsubsection{Metrics}
\label{sect:method_eval_metrics}

For evaluation, each outcome was treated as a binary classification task: given a patient’s history, the model provided an estimate of the probability of event $i$ occurring within $t$ days, and this was compared against the observed outcome. This allowed for the evaluation of single medical event prediction as well as bespoke phenotypes composed of multiple clinical codes or more complex definitions. Under this binary classification formulation, false positives and false negatives are quantified and correspond to the concepts of hallucinations and omissions in LLMs, respectively. Competing risks (such as death prior to the onset of an outcome) were not explicitly modelled as separate states; instead, the prediction of a death token terminated the sequence generation.

Discriminative performance was assessed using the area under the receiver operating characteristic curve (AUROC) \cite{roc} and the area under the precision-recall curve (AUPRC) \cite{average_precision}, the latter being commonly used under class imbalance \cite{pr_curve_w_imbalance_1, pr_curve_w_imbalance_2}. Overall performance and calibration were jointly measured using the Brier score \cite{brier_score}. We also generated ROC, precision-recall, and calibration curves for visual inspection. To reflect potential use in a clinical screening workflow, we also evaluated recall at a fixed 10\% false-positive rate (FPR10), consistent with prior studies \cite{fpr10} and quantifying performance at a hypothetical operational threshold.

Confidence intervals (95\%) were obtained via non-parametric patient-level bootstrapping with the percentile method, using 1,000 resamples for all analyses except the Phecode tasks (where 100 were used due to computational cost). 

\subsubsection{Subgroup Analyses}
\label{sect:method_subgroups}
We conducted subgroup analyses of mortality predictions at both the 30-day post–SARS-CoV-2–positive test and one-year horizons, stratifying test-set patients independently by age, sex, ethnicity, and vaccination status. All of these stratification variables were tokenised and included in the model's training data; age and vaccination status were embedded longitudinally within the patient timelines, while sex and ethnicity were included as static demographic tokens at the start of each timeline. For each subgroup, we calculated the AUROC and AUPRC in comparison with the overall cohort. Additionally, we investigated how performance on these metrics varied with the number of historical patient events provided as model input. Finally, we assessed changes in AUROC and AUPRC relative to the timing of the positive SARS-CoV-2 test to determine whether the Foresight-E model could model the changing epidemiology of COVID across the pandemic.

\subsection{Baseline Methods}
\label{section_baseline}

We benchmarked Foresight-E against supervised classifiers trained separately for all-cause mortality and hospitalisation at 30 days after a positive SARS-CoV-2 test, and for the same outcomes over a one-year horizon in 2023. 

These comparators used the same training split as Foresight-E but were task-specific, in contrast to Foresight-E’s zero-shot capability. The first was a simple logistic regression model \cite{log_regression} using age, sex, and ethnicity (one-hot encoded, meaning each categorical variable was represented as a binary vector where only the position corresponding to the observed category is set to one), with vaccination status added for acute SARS-CoV-2 outcomes. The second was an XGBoost model \cite{xg_boost} using count vectors of all medical codes from the Foresight-E vocabulary, providing equivalent structured clinical data but without temporal information. 

To allow training outcomes to be observed, input cut-offs were set to 1~December~2022 for the 30-day tasks and 1~January~2022 for the one-year tasks. This requirement highlights a key advantage of Foresight-E’s self-supervised learning: it is trained without requiring explicitly defined outcome labels. This eliminates the need to withhold a temporally separated subset of data for outcome labelling.
Performance for all models was evaluated using AUROC, AUPRC, and Brier score, as described in Section~\ref{sect:method_eval_metrics}.

\subsection{Governance}
\label{subsection_goverance}

The North East - Newcastle and North Tyneside 2 research ethics committee provided ethical approval for the CVD-COVID-UK/COVID-IMPACT research program (REC No 20/NE/0161) for approved research projects to access, within secure trusted research environments, whole-population, de-identified data from EHRs collected as part of patients’ routine healthcare.

The CVD-COVID-UK/COVID-IMPACT programme, led by the BHF Data Science Centre \cite{bhf_main}, received approval to access data in the NHSE SDE service for England from the Independent Group Advising on the Release of Data (IGARD) \cite{nhs_digital_agd} via an application made in the Data Access Request Service (DARS) Online system (ref. DARS-NIC-381078-Y9C5K) \cite{nhs_digital_data_access_request}. 

The CVD-COVID-UK/COVID-IMPACT Approvals \& Oversight Board that includes patient and public advisors \cite{bhf} subsequently granted approval to this project (CCU078: Foresight: a generative AI model of patient trajectories across the COVID-19 pandemic) in December 2023 to access the data within the NHSE SDE service for England.

Patient and public involvement was included in the approvals process and has continued to shape the research and communications through Patient and Public Involvement and Engagement sessions organised via British Heart Foundation (BHF) Data Science Centre \cite{bhf}.

\subsubsection{Data Availability}
\label{section_data_availability}
The data used in this study are available in the NHSE SDE service for England, but as restrictions apply, they are not publicly available \cite{nhs_sde}. 

The de-identified data used in this study were made available to accredited and approved researchers only. Those wishing to gain access to the data should contact \href{mailto:bhfdsc@hdruk.ac.uk}{bhfdsc@hdruk.ac.uk} in the first instance, noting that as detailed in Project Status \ref{section_status} data access is currently paused.

\subsubsection{Code Availability}
\label{section_code_availability}

All data preparation, model training, and evaluation code was intended to be released, following export and review via the SDE’s Safe Output Service, on GitHub at: \url{https://github.com/BHFDSC/CCU078_Foresight-England}, including a \texttt{requirements.txt} specifying all package versions. However as detailed in Project Status \ref{section_status} data access is currently paused meaning code cannot be exported and shared.

The code used to produce the placeholder results figures was developed outside the SDE, without access to the underlying data, and is available at \url{https://github.com/simonEllershaw/foresight_placeholder_graphs}

Due to data restrictions, trained model weights and artefacts are only accessible to a subset of approved CVD-COVID-UK/COVID-IMPACT consortium researchers on a dedicated Foresight-E cluster within the NHSE SDE. Those wishing to gain access to the data should contact \href{mailto:bhfdsc@hdruk.ac.uk}{bhfdsc@hdruk.ac.uk} in the first instance, noting that as detailed in Project Status \ref{section_status} data access (including model weights, artefacts and code) is currently paused.

All analyses were executed within the NHSE SDE \cite{nhs_sde} using Databricks Runtime 14.3 LTS for ML \cite{databricks_14_3_ml}; training/evaluation used an AWS g5.48xlarge instance with eight NVIDIA A10 GPUs \cite{g5_cluster, nvidia_a10}. AWS and Databricks had no access to the underlying datasets or trained AI model and no control over the research or its findings.

\newpage
\section{Results}
\label{sect:results}

As detailed in Project Status \ref{section_status}, an initial quantitative evaluation has been completed, but the ongoing pause in data access prevents export of the results from the SDE. To be transparent about what was evaluated and what outputs were intended for publication, we present the relevant tables and figures with placeholder values, marked with '$\_$', and placeholder figures where data cannot be shown.

\subsection{30-day COVID-19 Mortality and Hospitalisation Prediction}
\label{subsection_results_covid}

To demonstrate the utility of Foresight-E in predicting the direct effects of COVID-19 without fine-tuning, we applied it to the important clinical prediction task of forecasting 30-day mortality and hospitalisation following a positive SARS-CoV-2 test.

\begin{figure}[!h]
    \centering
    \begin{subfigure}{\linewidth}
        \centering
        \includegraphics[width=0.7\linewidth]{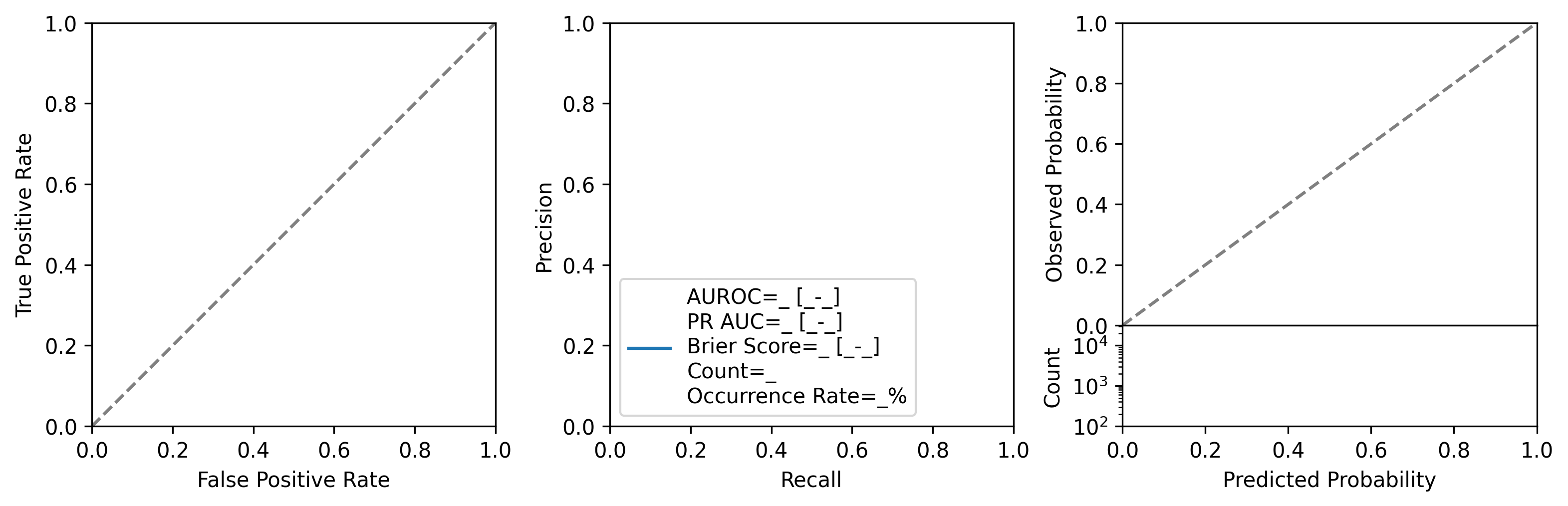}
        \caption{Mortality}
        \label{fig_covid_death}
    \end{subfigure}

    \begin{subfigure}{\linewidth}
        \centering
        \includegraphics[width=0.7\linewidth]{figures/fig_performance_curve.png}
        \caption{Hospitalisation}
        \label{fig_covid_hosp}
    \end{subfigure}
    \caption{\textbf{PLACEHOLDER.} ROC, PR, and calibration curves showing Foresight-E's prediction performance for 30-day mortality and hospitalisation following a positive SARS-CoV-2 test. Shaded regions represent 95\% confidence intervals estimated via 1{,}000 bootstrap iterations. Counts are rounded to the nearest five to comply with the NHSE SDE's disclosure control requirements.}
    \label{fig_30_day_covid}
\end{figure}

Foresight-E achieved an AUROC of $\_$ ($\_$-$\_$, 95\% CI) and $\_$ ($\_$-$\_$, 95\% CI) for 30-day prediction of mortality and hospitalisation, respectively. For comparison, the baseline supervised logistic regression and XGBoost models yielded AUROCs of $\_$ and $\_$, and $\_$ and $\_$, respectively. Calibration and discrimination was quantified using Brier scores, yielding $\_$ and $\_$ for the mortality and hospitalisation endpoints, respectively. At a 10\% false positive rate (FPR), detection rate (DR10) was $\_$ and $\_$, respectively. ROC, precision-recall, and calibration curves for each task are shown in Fig.~\ref{fig_30_day_covid}, with 95\% confidence intervals estimated via 1,000 sampled bootstrap iterations.

\newpage

\subsubsection{COVID-19 Outcomes Across the Pandemic}

To evaluate Foresight's ability to model the changing epidemiology of COVID-19 across the duration of the pandemic (including varying case rates, mortality, and circulating viral variants), we performed a subgroup analysis stratified by the calendar year and month of the patient’s first positive SARS-CoV-2 test. Across these temporal strata, we observed AUROC values ranging between $\_$ and $\_$. On the 2023 data, a temporal holdout set excluded from training or validation, the computed AUROC was $\_$.

\begin{figure}[!h]
\centering
\includegraphics[width=0.6\linewidth]{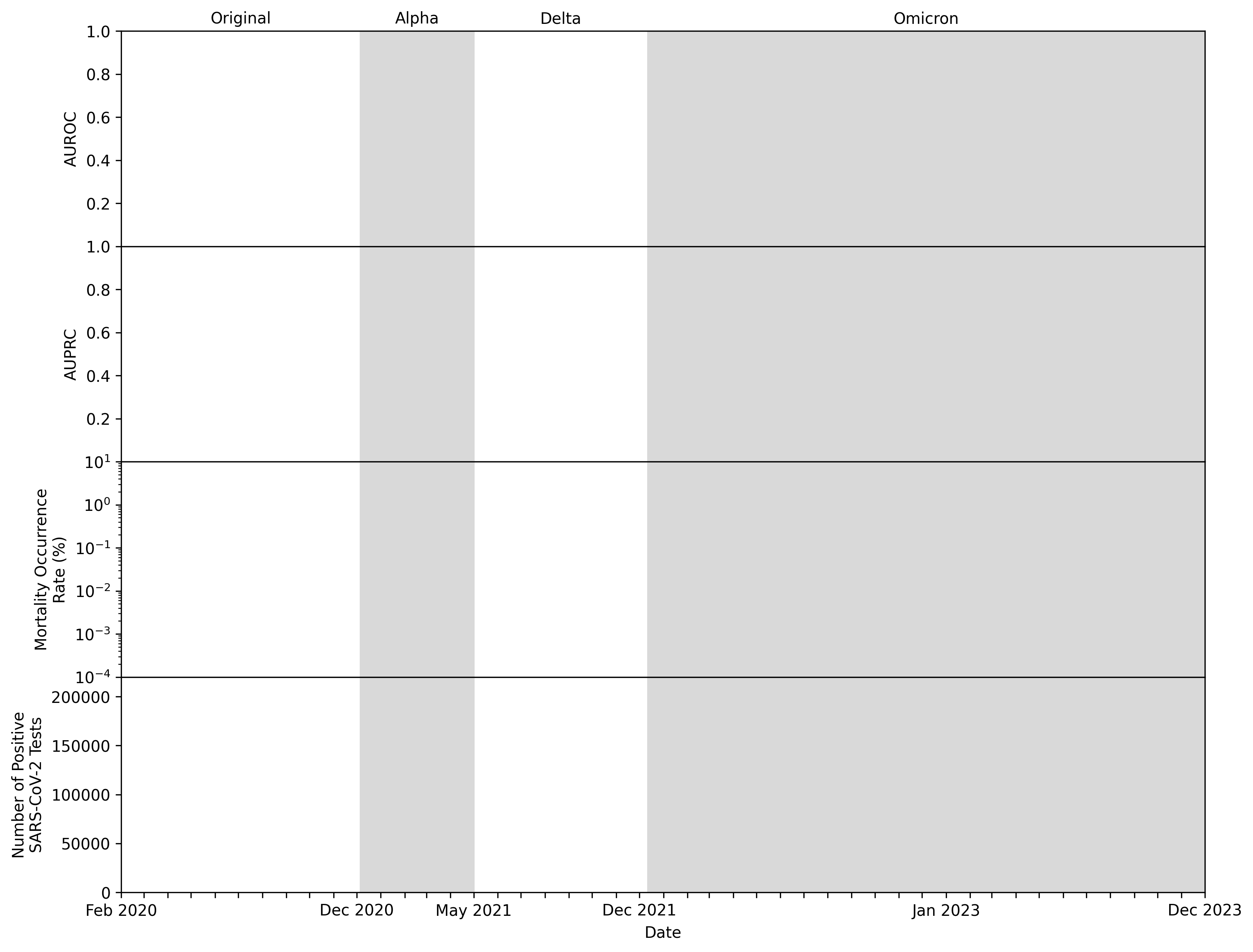}
\caption{\textbf{PLACEHOLDER.} Temporal evaluation of 30-day mortality prediction across the COVID-19 pandemic. Monthly performance of Foresight-E in predicting death within 30 days of a positive SARS-CoV-2 test, measured by AUROC (top panel) and AUPRC (second panel). Error bars represent $95\%$ confidence intervals estimated via 1{,}000 bootstrap iterations. Also shown are monthly mortality occurrence rates (third panel) and the number of positive SARS-CoV-2 tests (bottom panel), illustrating changes in disease dynamics over time. Shaded background regions denote the predominant SARS-CoV-2 variants \cite{wilde2023hospital} with labels shown along the upper x-axis. Counts are rounded to the nearest five to comply with the NHSE SDE's disclosure control requirements.}\label{fig_covid_timeline}
\end{figure}

\newpage

\subsubsection{COVID-19 Outcomes by Subgroup}

\begin{figure}[!h]
\centering
\includegraphics[width=0.6\linewidth]{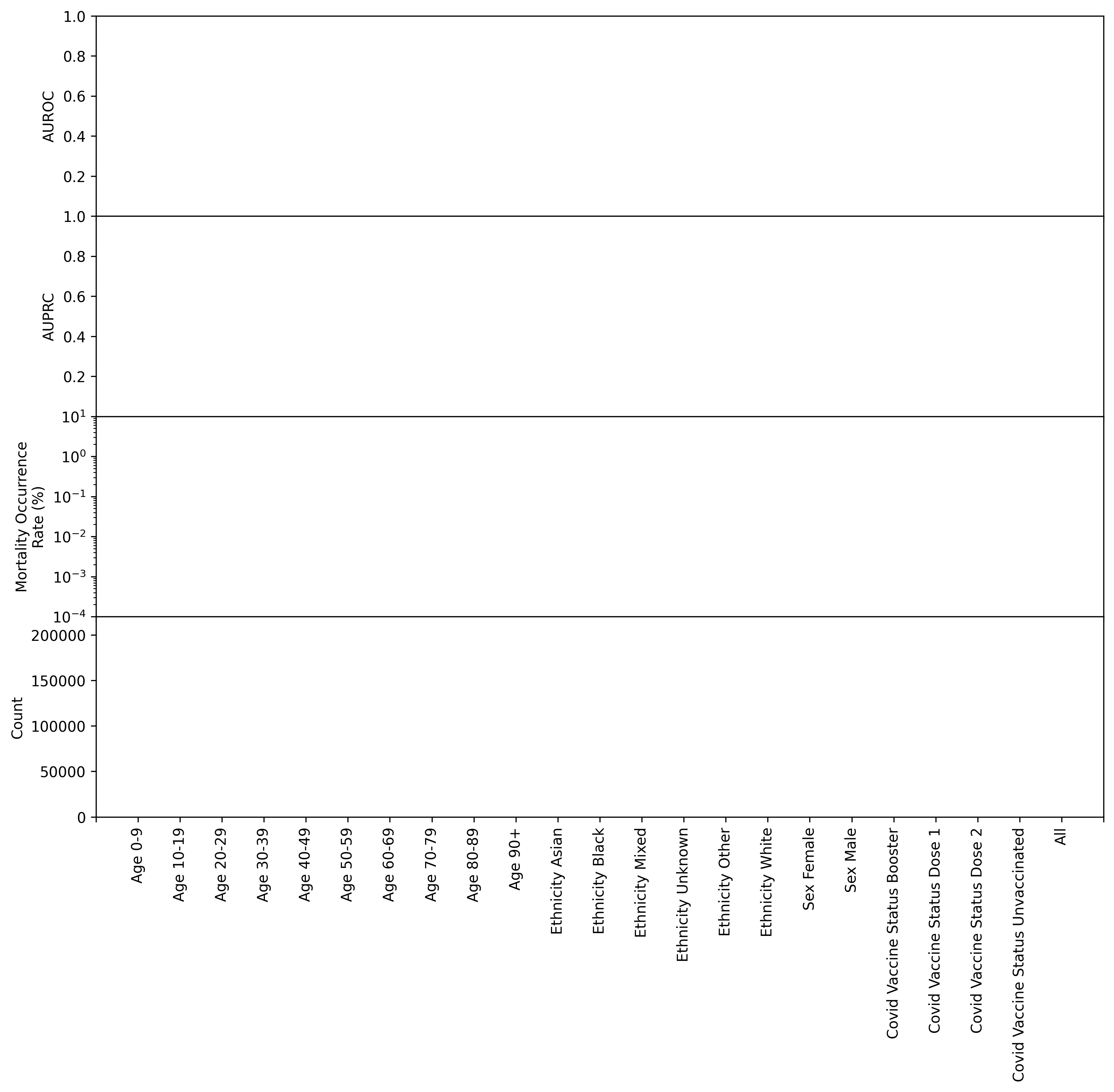}
\caption{\textbf{PLACEHOLDER.} Variation in AUROC and AUPRC metrics for the Foresight-E model predicting 30-day mortality after a positive SARS-CoV-2 test, stratified by age, ethnicity, sex, and COVID-19 vaccination status. Error bars show 95\% confidence intervals calculated by 1{,}000 bootstrap iterations. Also shown is the mortality occurrence rate, as well as the count of patients for each group. Performance is consistent across ethnicity and sex, but varies notably by age group and vaccination status. Counts are rounded to the nearest five to comply with the NHSE SDE's disclosure control requirements.}
\label{fig_covid_sub_group}
\end{figure}

Fig.~\ref{fig_covid_sub_group} shows the performance of the Foresight-E model in predicting 30-day mortality following a positive SARS-CoV-2 test, stratified by age, ethnicity, sex, and COVID-19 vaccination status. 

To assess for performance disparities across demographic groups, we computed AUROC scores of $\_$ for White patients, $\_$ for Asian patients, $\_$ for Black patients, and $\_$ for Mixed/Other ethnicities. Between sexes, AUROC scores were $\_$ for male patients and $\_$ for female patients. Across age brackets, AUROC values measured $\_$ to $\_$, mapped against a crude mortality rate of $\_$\% to $\_$\% in those respective demographics.

\newpage

\subsubsection{COVID-19 Outcomes By Event History}
\begin{figure}[!h]
\centering
\includegraphics[width=0.5\linewidth]{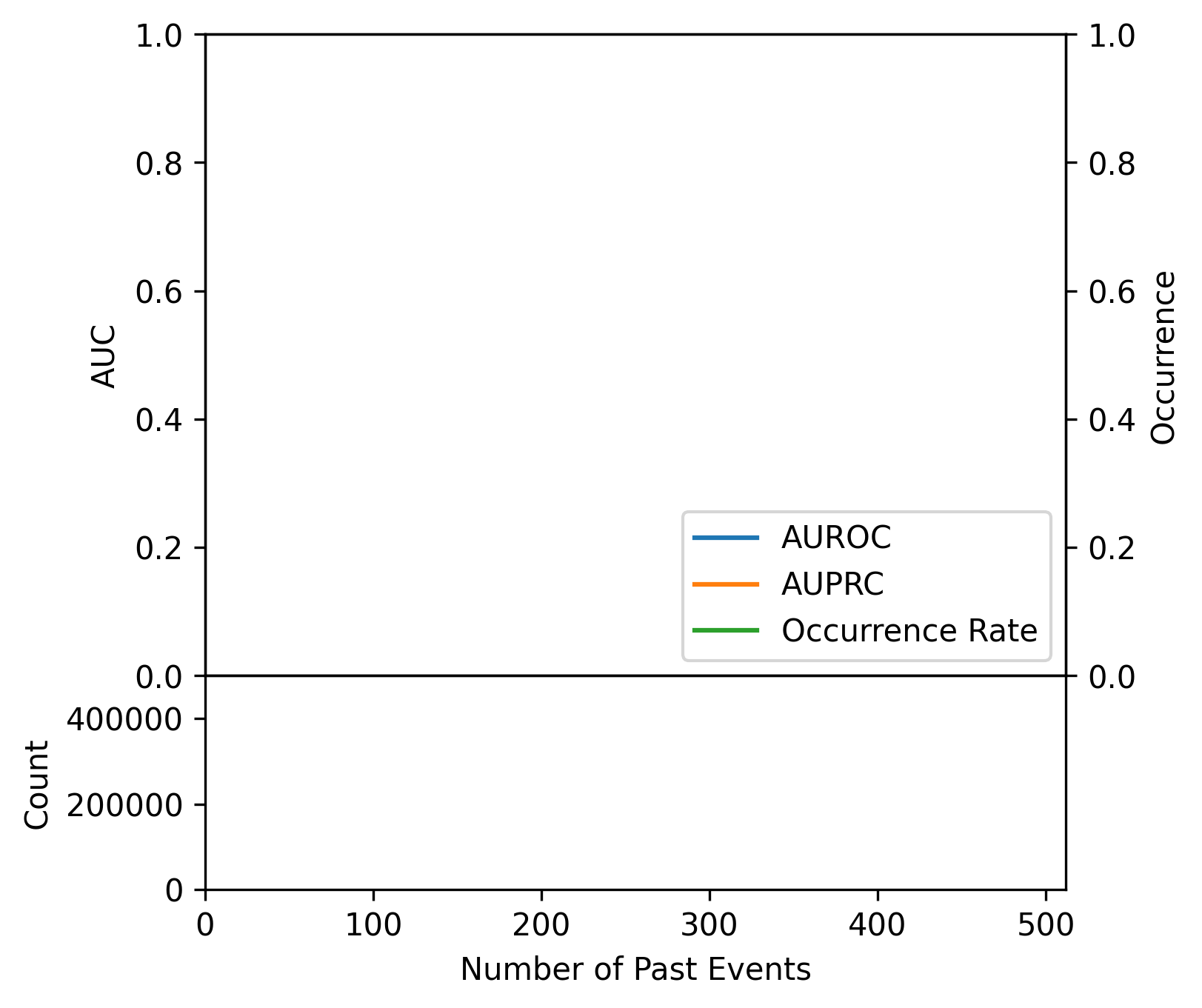}
\caption{\textbf{PLACEHOLDER.} Variation in AUROC and AUPRC metrics for the Foresight-E model predicting 30-day mortality after a positive SARS-CoV-2 test with the number of past medical events recorded in a patient’s timeline. Also shown are the binned frequency of past events and corresponding mortality occurrence rates. A bin size of 16 events was used, and the final bin includes all patients with more than 498 events. Counts are rounded to the nearest five to comply with the NHSE SDE's disclosure control requirements.}\label{fig_covid_num_past_events}
\end{figure}

Fig.~\ref{fig_covid_num_past_events} plots model performance predicting 30-day mortality after a positive SARS-CoV-2 test against the number of historical patient events provided as input to Foresight-E. We measured a correlation of $\_$ between the number of past medical events and AUROC, contextualised by a correlation of $\_$ between the number of past medical events and the outcome. The distribution of the number of past events across the cohort exhibited a skewness of $\_$.

\newpage

\subsection{Predicting Indirect Effects of COVID-19 in 2023}
\label{subsection_results_1_year}

To evaluate Foresight-E's ability to predict the indirect effects of the COVID-19 pandemic, we simulated a prospective deployment. All patient timelines were truncated at 1~January~2023, using the \texttt{YEAR\_START} token. Foresight-E was not exposed to any event data from 2023 during training, thereby testing generalisation to unseen future events. 

\subsubsection{Emergency Hospitalisation and Mortality Prediction}

\begin{figure}[!h]
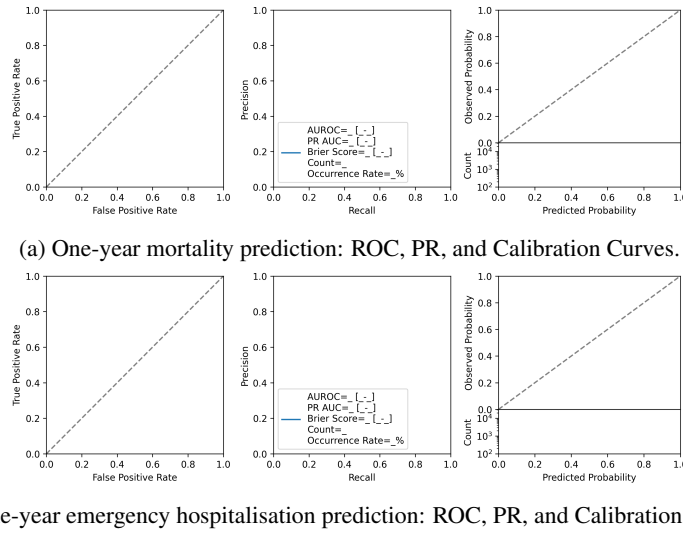

    \centering
    \begin{subfigure}{\linewidth}
        \centering
        \includegraphics[width=0.65\linewidth]{figures/fig_performance_curve.png}
        \caption{One-year mortality prediction: ROC, PR, and Calibration Curves.}
        \label{fig_1_year_death}
    \end{subfigure}

    \begin{subfigure}{\linewidth}
        \centering
        \includegraphics[width=0.65\linewidth]{figures/fig_performance_curve.png}
        \caption{One-year emergency hospitalisation prediction: ROC, PR, and Calibration Curves.}
        \label{fig_1_year_hosp}
    \end{subfigure}
    \caption{\textbf{PLACEHOLDER.} Prediction performance of Foresight-E in 2023 using ROC, PR, and calibration curves. Shaded regions represent 95\% confidence intervals estimated via 1{,}000 bootstrap iterations. Counts are rounded to the nearest five to comply with the NHSE SDE's disclosure control requirements.}
    \label{fig:1_year_curves}
\end{figure}

The ROC, precision-recall, and calibration curves are presented in Fig.~\ref{fig:1_year_curves}. 

Foresight-E achieved an AUROC of $\_$ ($\_$-$\_$, 95\% CI) and $\_$ ($\_$–$\_$, 95\% CI) for emergency hospitalisation and mortality, respectively. For comparison, the baseline supervised logistic regression and XGBoost models yielded AUROCs of $\_$ and $\_$, and $\_$ and $\_$, respectively. Calibration and discrimination was quantified using Brier scores, yielding $\_$ and $\_$ for emergency hospitalisation and mortality endpoints, respectively.

\subsubsection{Phecode Predictions}

\begin{figure}[!h]
\centering
\includegraphics[width=0.4\linewidth]{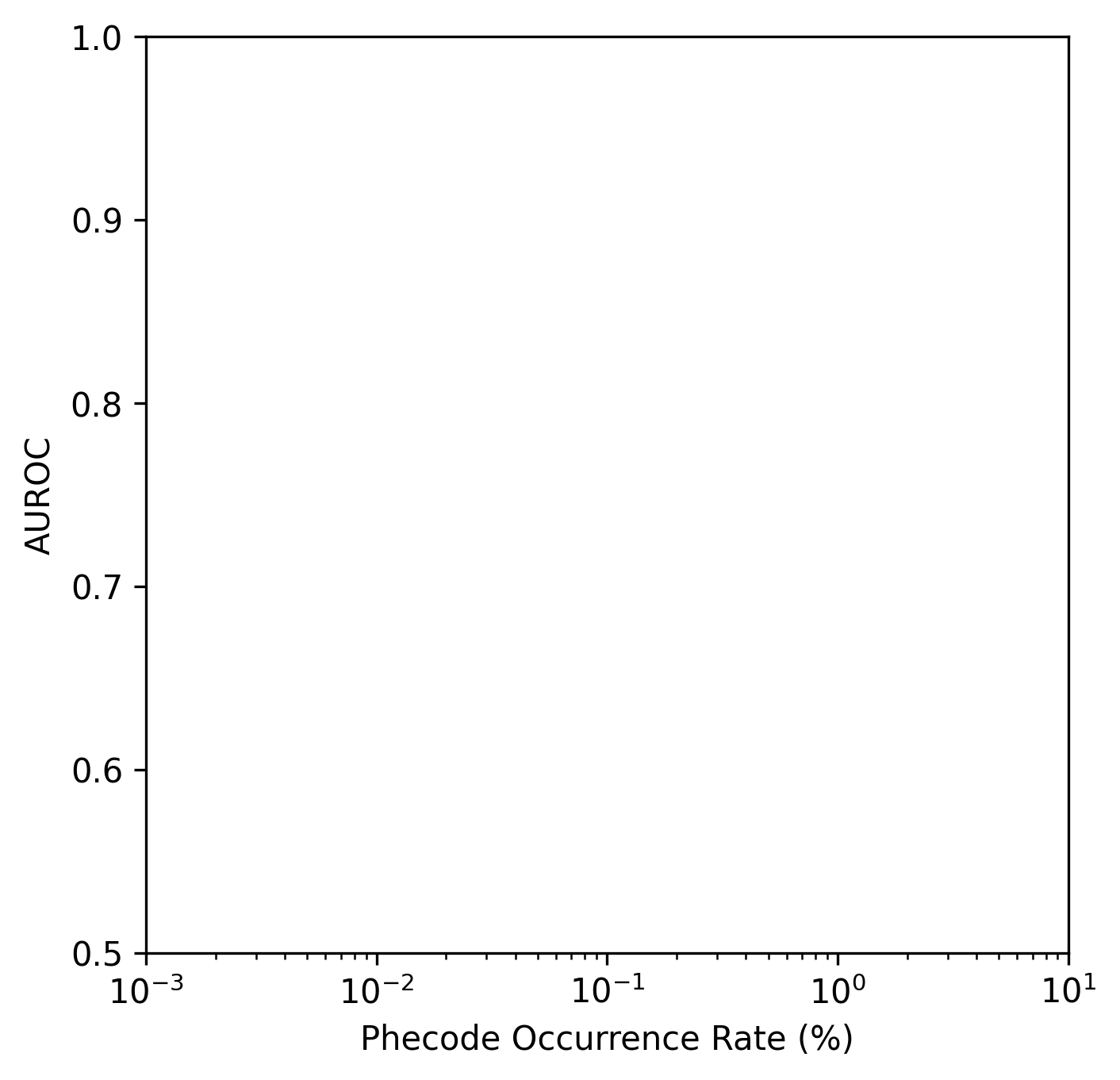}
\caption{\textbf{PLACEHOLDER.} Relationship between AUROC and event occurrence rate for Foresight-E’s 1-year prediction in 2023. $\_$ AUROCs were observed for more $\_$ Phecodes, such as $\_$.}\label{fig_1_year_phecodes}
\end{figure}

To assess the breadth of Foresight-E's capacity to predict the indirect effects of COVID, we evaluated performance on over 1,400 distinct Phecodes, previously validated groupings of ICD codes representing clinically meaningful phenotypes \cite{phecodes}. AUROC values varied $\_$-$\_$ across Phecodes, with an $\_$ association between event prevalence and predictive performance (Fig.~\ref{fig_1_year_phecodes}). 

\newpage
\section{Discussion}
\label{sect:diss}

We developed Foresight-E, a 243-million-parameter generative foundation model trained de novo on longitudinal electronic health records from 54.9 million patients, entirely within the NHS England Secure Data Environment. Designed as a research pilot for COVID-19, the model was evaluated on its zero-shot capability to predict both the direct acute outcomes of SARS-CoV-2 infection and the pandemic's enduring, system-wide indirect effects. By successfully engineering this pipeline within existing national infrastructure, this work establishes a methodological template for future pandemic response and broader applications, from clinical risk prediction to population-level healthcare planning. As quantitative results on our 6.1-million-patient test set are currently withheld pending an ongoing governance review, we focus our discussion below on the core methodological strengths and limitations of this work, including data integration, modelling and evaluation, before discussing future directions for the field.

\subsection{Data}
\label{sect:dis_data}

Foresight-E’s key strength is its first use of national-scale, routinely collected EHRs spanning primary care, secondary care, COVID-19, and death registrations for the development of a generative AI model. This ensures Foresight-E is trained on diverse data representative of the general population \cite{pineda2024ethnicity}, helping to mitigate algorithmic bias \cite{arora2023}, enabling prediction of rare events \cite{thygesen2025rare,lancet_rheum}, and generating evidence for the methodology’s translational potential through aligning training data with intended use populations. Given that primary care accounts for most healthcare delivery, integrating this data provides a more complete representation of an individual’s health, enables earlier risk stratification, and may ultimately guide preventive interventions to avert disease onset, complications and costly secondary care.

Despite its breadth, the datasets reflect typical challenges of routinely collected health data: incomplete or imprecise coding \cite{kostopoulou2021}, historical biases in care \cite{sjoding2020spo2, arora2023}, and shifts in recording practices \cite{zghebi2022clinical}, especially during the COVID-19 pandemic \cite{pineda2025disparity}. Notably, the GDPPR primary care dataset contains a subset of codes, specifically, those already available from previous GPES extracts that were deemed relevant for COVID-related research. As a result, it omits examples of both common and rare diseases, as well as signs and symptoms data that is crucial for understanding disease presentation and evolution - for example, COVID presenting with anosmia, or non-specific symptoms of long COVID, such as fatigue. 

Furthermore, GDPPR only contains those alive on or after 1~November~2019. This means that individuals who died before this date are excluded, restricting the ability to model or evaluate long-term disease progression or events in the pre-COVID era. While this is not a limitation for models focused on COVID-19 infection, since such cases did not exist before early 2020, it does reduce the available historical context. Expanding the temporal window to include earlier records would be expected to increase predictive performance and improve the capacity to model long-term outcomes, but would require methods capable of efficiently handling longer sequences \cite{context_clues} and risk issues such as immortal time-bias without changes in dataset inclusion criteria.

While Foresight-E is strictly confined to the structured data approved for COVID-19 research, future applications of this methodological approach could be enhanced by incorporating additional modalities such as clinical notes or medical imaging. The development of foundational multi-modal transformer models in the general domain shows how the methods presented here could be extended \cite{clip, perceiver}. However, such data is not currently available at a national scale, and provisioning new data streams for future models would require appropriate governance, infrastructure, and technical methods, including de-identification and pre-processing.

\subsection{Model}

Foresight-E uses a 243-million-parameter Llama 2-style transformer decoder. While larger than most prior EHR models \cite{delphi} and trained on national scale data, the scale of data, model size, and compute is still orders of magnitude smaller than the general domain \cite{llama_4}.

Foresight-E is designed to be data-driven, rather than relying on predefined parametric models such as exponential hazard–style structures \cite{delphi, motor}. Although we trained from scratch due to a custom vocabulary and NHSE SDE import restrictions, smaller-scale studies indicate that starting with an LLM pre-trained on vast general data and then fine-tuning for medical-event prediction is beneficial to performance \cite{foresight_2, llms_are_ehr_encoders}. If future NHSE SDE policy permits importing pretrained models, initialising from a general model, and adapting in-domain is a promising direction.

Our tokenisation strategy preserves the recorded granularity of clinical codes and day-level timing of the EHR data , rather than aggregating codes into higher level categories \cite{delphi} or phenotypes \cite{behrt}, filtering low-occurrence events \cite{foresight} or binning time \cite{ethos}. This maximised diagnostic specificity and allowed prediction of rare events often excluded in other models \cite{lancet_rheum}. However, this enlarged the vocabulary and prediction space and reduced per-token training frequency. 

Furthermore, all event codes present in the training set were added to the tokeniser’s vocabulary. Therefore, during inference, codes absent from the training data could not be represented because the vocabulary did not include an unknown token (e.g., \texttt{<UNK>}), so we excluded them, discarding potentially clinically relevant information. The tokeniser could be further improved by exploiting the hierarchical structure of clinical ontologies \cite{ethos}, for which there is evidence of performance gains on EHR prediction tasks \cite{medtok}.

Building on previous work that models chronological age and relative time between events, we additionally encoded absolute calendar time to model the changing healthcare system during the COVID pandemic. Because Foresight-E has a maximum context length of 1,024 tokens, long histories required truncation. Naïve token-level truncation (e.g., dropping the first $n$ tokens) corrupts timelines by removing demographic, year-start, or age tokens or invalidating time-difference tokens. We therefore applied event-level truncation followed by re-tokenisation as a pre-processing step for training. During autoregressive inference, however, all operations must run on the GPU, making event-level truncation infeasible. We therefore used fixed maximum input and generation lengths, which occasionally discarded more tokens than necessary and, in some cases, prevented forecasted trajectories from reaching the final time horizon (e.g. 1 year).

Previous work \cite{ethos} constructs training examples by concatenating all patient EHRs into a single sequence and then randomly sampling subsequences of a fixed length. Although this maximises computational efficiency by eliminating the need for padding tokens, it allows individual training examples to span across multiple patients. This risks the model learning spurious cross-patient relationships. To avoid this, we adopt a patient boundary-preserving strategy \cite{ehr_scaling}, forming sequences strictly at the patient level, which comes at the expense of computational efficiency. Future work could incorporate intra-document causal masking \cite{document_masking} to enable efficient concatenation without information leakage.
\subsection{Inference}

We extended prior probabilistic patient-trajectory approaches \cite{ethos}, applying Foresight-E in a zero-shot setting to forecast over 1,400 medical events, across 30-day and 1-year time horizons relevant to direct and indirect COVID-19 outcomes. This offers the broad ability to predict differential diagnoses and clinical trajectories, rather than single outcomes, but is computationally intensive, particularly at a population scale. Comparative studies with task-specific fine-tuning are needed to clarify efficiency–flexibility trade-offs.

Our current setup generates 48 trajectories per patient, constrained practically by the NVIDIA A10 GPU's memory. Exploring how probability estimates stabilise with the number of samples and alternative decoding strategies (beam search, top-k, temperature) may yield gains in efficiency, accuracy, and calibration, but would require tuning. Furthermore, quantifying the temporal horizon over which Foresight-E can reliably forecast patient outcomes, both in absolute time and in terms of tokens, requires further study.

\subsection{Evaluation}

We evaluated Foresight-E's ability to predict COVID-19 outcomes in two settings designed to reflect real-world deployment challenges.

First, we assessed its ability to predict direct COVID-19 outcomes during the pandemic, a period marked by shifting conditions such as emerging viral variants, changing testing protocols, public health interventions and population immunity. Unlike earlier models \cite{ethos}, Foresight-E explicitly encodes both absolute calendar time and patient age, enabling it to adapt to these temporal shifts. Forecasting future novel threats, outside the current training data and vocabulary, could be supported by continued pretraining with vocabulary expansion on the latest batches of newly collected data.

Second, we sought to predict the indirect effects of COVID-19 and simulate a prospective deployment by predicting events in 2023, one year beyond the training period, on over 1,400 Phecodes, emergency hospitalisation, and all-cause mortality. COVID-19 highlights the challenges of temporal data shifts. Foresight-E was trained during the height of the COVID-19 pandemic, a period of profound healthcare system disruption and excess all-cause mortality, which left an enduring and evolving legacy in the evaluation period of 2023, encompassing the indirect effects the pandemic exerted on individuals, healthcare systems, and society at large \cite{indirect_covid_2}.

This represents, to our knowledge, the broadest zero-shot evaluation of an EHR foundation model to date. Working at the scale of the English population meant facing a low occurrence rate for many acute outcomes, in contrast to models trained solely on high-acuity inpatient cohorts (e.g., MIMIC-IV \cite{mimic_iv}). While this sparsity poses challenges, it also demonstrates Foresight-E’s potential utility for COVID-specific population screening as well as high-risk patient monitoring, such as identifying vulnerable individuals to prioritise targeted interventions (e.g. vaccination, antivirals), by training on both general ‘healthy’ and acutely ill patient timelines. In order to critically evaluate Foresight-E’s potential for population screening, we additionally calculated recall at a fixed 10\% false-positive rate, consistent with prior studies \cite{fpr10} and representing a hypothetical operational threshold.

Whilst evaluating each patient trajectory as multiple binary prediction tasks allowed use of established metrics (AUROC, AUPRC, Brier score) to quantify performance on clinically-relevant tasks, these fail to capture the overall fidelity of generated trajectories, an important direction for future work.

A key limitation of this work is the inability to benchmark against commonly used clinical risk prediction tools, due to a lack of required data (e.g. 4C Mortality Score for COVID requires physiological measurements and test results \cite{knight2020risk}), limited data duration (e.g. QRisk measures 10-year cardiovascular disease risk \cite{qrisk}) and the fact that where risk scores were recorded the resulting outcomes were then conditioned on resultant clinical decision making \cite{logan2025early}. For methodological comparators NHSE SDE constraints meant external pretrained models could not be imported and resource limitations restricted the number of comparator models trained. 

External validation, to assess model generalisability across different populations, was not possible as Foresight-E was developed entirely inside the NHSE SDE and the trained model could not be transferred out of the environment (including to another SDE), nor could other cohorts be imported into the NHSE SDE.

Explainability techniques, such as attention-weighted visualisation, gradient-based saliency, or counterfactual generation, are left as directions for future work but could help clinicians understand the model’s forecasts by scrutinising learned associations for known or plausible patterns, as well as the presence of spurious correlations, such as shortcut learning. Such methods could support safe adoption in practice by explaining why the model anticipates particular outcomes, as well as identifying potentially modifiable risk factors as targets for interventions to optimise health.

\subsection{Future Directions and Considerations}
\label{sect:diss_use_cases}

Foresight-E was developed as a research pilot strictly for COVID-19-related research and is not a validated clinical tool. The model is therefore confined to this scope, and any future directions for the methodology are entirely contingent on navigating the significant challenges outlined below.

The generative, zero-shot forecasting methodology demonstrated by models like Foresight-E offers broad potential, including forecasting population health, stratifying groups at increased risk of adverse outcomes, and enabling personalised risk prediction to guide preventive interventions. Beyond direct clinical care, potential applications include improving clinical trial efficiency through prognostic enrichment or advancing drug discovery by better modelling disease trajectories. Models like Foresight are by design associative, rather than causal, and therefore a priority area for research is the robust evaluation of the extent to which counterfactual questions can be answered, a crucial step toward creating robust digital twins and enabling trustworthy in-silico trials.

Moving from research to application would require secure, real-time model deployment, a capability beyond the current NHSE SDE infrastructure. This gap is particularly critical for large generative models, which can inadvertently memorise and expose sensitive training data \cite{carlini2021extracting}. A potential mitigation strategy is to deploy such models within a secure environment behind narrowly scoped APIs. These would provide only predefined, validated outcomes, such as calibrated risk scores, rather than open-ended generative trajectories, thereby constraining vectors for data extraction and mitigating privacy risks \cite{grainmatter}.

However, these ambitions are secondary to the fundamental legal and governance challenges. The most significant barrier to any extension of this methodology is that there is currently no lawful basis to use this national dataset beyond emergency directions issued specifically for COVID-19 pandemic research \cite{copi}. Therefore, advancing this work requires new legal permissions through transparent public consultation, paired with a clear public benefit. A lawful basis and a social licence must go hand in hand; this requires deep and sustained engagement with patients, the public, and professional bodies. Furthermore, the path from a research model to a trustworthy clinical tool would necessitate additional rigorous evaluation and a clear route to regulatory approval as a medical device. Addressing these socio-technical challenges is the central prerequisite for future progress in this domain.

\newpage
\section{Conclusion}

We have presented Foresight-E, a 243-million-parameter transformer trained on national-scale EHR data from 54.9 million NHS patients and evaluated on its ability to perform zero-shot prediction across $\sim$1.4k COVID-related outcomes for a 6.1-million-patient test set. We outline the data integration pipeline, tokenisation strategy, model architecture, training procedure, and inference and evaluation framework.

Although quantitative results are currently withheld pending ongoing discussions, this work demonstrates that it is technically feasible to develop a foundation model for healthcare entirely within existing NHS infrastructure. By combining routinely collected population-scale EHRs with modern generative modelling, Foresight-E offers a blueprint for zero-shot healthcare AI systems.

Rebuilding beyond COVID-restricted datasets, expanding access to broader clinical modalities, and developing safe deployment pathways could enable models like Foresight-E to support both population-level planning and individualised care. Realising this potential will require not only technical advances but also transparent governance, sustained public and professional engagement, and rigorous evaluation in real-world clinical settings to generate evidence for regulatory approval.

\newpage
\section{Acknowledgements}
\label{section_acknowledgements}

The British Heart Foundation Data Science Centre (grant No SP/19/3/34678, awarded to Health Data Research UK) funded co-development (with NHS England) of the SDE service for England, provision of linked datasets, data access, user software licences, computational usage, and data management and wrangling support, with additional contributions from the HDR UK Data and Connectivity component of the UK Government Chief Scientific Adviser’s National Core Studies program to coordinate national COVID-19 priority research. Consortium partner organisations funded the time of contributing data analysts, biostatisticians, epidemiologists, and clinicians.

AWS provided the compute credits which made this work possible. Databricks provided technical support. AWS and Databricks had no access to the underlying datasets or trained AI model and no control over the research or its findings.


SE and CT are funded by the UKRI Centre for Doctoral Training in AI-enabled healthcare (EP/S021612/1). CT also receives support from a King’s College London AI+ Senior Academic Fellowship, a MRC Clinical Top-Up, NIHR Biomedical Research Centre at UCL Hospital NHS Trust, and Health Data Research UK. 

AMW is supported by the BHF Data Science Centre (HDRUK2023.0239), Health Data Research UK (Big Data for Complex Disease-HDR-23012), and as an NIHR Research Professor (NIHR303137). Her research is also supported by core funding from the British Heart Foundation (RG/F/23/110103), NIHR Cambridge Biomedical Research Centre (NIHR203312) [*], BHF Chair Award (CH/12/2/29428), Cambridge BHF Centre of Research Excellence (RE/24/130011), and by Health Data Research UK (HDRUK2023.0028), which is funded by the UK Medical Research Council, Engineering and Physical Sciences Research Council, Economic and Social Research Council, Department of Health and Social Care (England), Chief Scientist Office of the Scottish Government Health and Social Care Directorates, Health and Social Care Research and Development Division (Welsh Government), Public Health Agency (Northern Ireland), British Heart Foundation and the Wellcome Trust.

RD’s work is supported by (1) National Institute for Health Research (NIHR) Biomedical Research Centre at South London and Maudsley NHS Foundation Trust and King’s College London. (2) Health Data Research UK, which is funded by the UK Medical Research Council, Engineering and Physical Sciences Research Council, Economic and Social Research Council, Department of Health and Social Care (England), Chief Scientist Office of the Scottish Government Health and Social Care Directorates, Health and Social Care Research and Development Division (Welsh Government), Public Health Agency (Northern Ireland), British Heart Foundation and Wellcome Trust. (3) The National Institute for Health Research University College London Hospitals Biomedical Research Centre.


This work was carried out with the support of the BHF Data Science Centre led by HDR UK (BHF Grant no. SP/19/3/34678). This study made use of de-identified data held in the NHSE SDE service for England and made available via the BHF Data Science Centre’s CVD-COVID-UK/COVID-IMPACT consortium. This work used data provided by patients and collected by the NHS as part of their care and support. We would also like to acknowledge all data providers who make health-relevant data available for research.

We thank partners at NHS England, AWS, Databricks, and the BHF Data Science Centre for their support, without which this project would not have been possible. We extend our sincere gratitude to the public contributors of the British Heart Foundation Data Science Centre for providing critical review, stimulating discussion, and constructive feedback throughout the project's lifecycle, including their invaluable guidance in shaping the lay summary.

\section{Contributor Statement}
\label{section_contributions}

S.E. and C.T. contributed equally to this work and share joint first authorship.

SE: Conceptualisation, Data curation, Formal analysis, Investigation, Methodology, Software, Validation, Visualisation, Writing – original draft, Writing – review \& editing.

CT: Conceptualisation, Data curation, Formal analysis, Funding acquisition, Investigation, Methodology, Project administration, Resources, Software, Supervision, Validation, Visualisation, Writing – original draft, Writing – review \& editing.

ZK: Methodology, Writing – review \& editing.

SD: Methodology, Writing – review \& editing.

HH: Writing – review \& editing.

CS: Funding acquisition, Resources, Writing – review \& editing.

AMW: Writing – review \& editing.

AS: Writing – review \& editing.

RD: Conceptualisation, Funding acquisition, Methodology, Resources, Supervision, Writing – review \& editing.

In strict accordance with the information governance and data security protocols of the NHS England Secure Data Environment (SDE), access to the underlying datasets and dedicated Foresight cluster was restricted to only CT and SE. Consequently, all data curation, model training, formal data analysis, software implementation, and quantitative evaluation were conducted exclusively by CT and SE within the secure environment. CT is the guarantor for this work.

CS was previously the Director of the British Heart Foundation (BHF) Data Science Centre and coordinated approvals for and access to data within the NHS Digital Trusted Research Environment for England for CVD-COVID-UK/COVID-IMPACT.

\section{Disclosures}
\label{section_disclosures}

SE contracted part-time for Parexel International during the period this work was conducted.

CT was previously employed by LifeArc, and has received research funding via the UCL-GSK Phenomics Hub from GSK. 

ZK is a co-founder of Nuraxi. 

RD is a co-founder of CogStack and Onsentia.

ADS receives research funding from BMJ Publishing Group.

The remaining authors declare no competing interests.

None of these commercial organisations had any involvement in the funding, study design, data access, model training, evaluation, or execution of the Foresight-England project, nor do they have access to the underlying data, code, or model weights.

\newpage
\bibliography{bibliography}

@article{gpt3,
  title={Language models are few-shot learners},
  author={Brown, Tom and Mann, Benjamin and Ryder, Nick and Subbiah, Melanie and Kaplan, Jared D and Dhariwal, Prafulla and Neelakantan, Arvind and Shyam, Pranav and Sastry, Girish and Askell, Amanda and others},
  journal={Advances in neural information processing systems},
  volume={33},
  pages={1877--1901},
  year={2020}
}

@article{attention_is_all_you_need,
  title={Attention is all you need},
  author={Vaswani, Ashish and Shazeer, Noam and Parmar, Niki and Uszkoreit, Jakob and Jones, Llion and Gomez, Aidan N and Kaiser, {\L}ukasz and Polosukhin, Illia},
  journal={Advances in neural information processing systems},
  volume={30},
  year={2017}
}

@article{foresight,
  title={Foresight—a generative pretrained transformer for modelling of patient timelines using electronic health records: a retrospective modelling study},
  author={Kraljevic, Zeljko and Bean, Dan and Shek, Anthony and Bendayan, Rebecca and Hemingway, Harry and Yeung, Joshua Au and Deng, Alexander and Baston, Alfred and Ross, Jack and Idowu, Esther and others},
  journal={The Lancet Digital Health},
  volume={6},
  number={4},
  pages={e281--e290},
  year={2024},
  publisher={Elsevier}
}

@article{foresight_2,
  title={Large Language Models for Medical Forecasting--Foresight 2},
  author={Kraljevic, Zeljko and Yeung, Joshua Au and Bean, Daniel and Teo, James and Dobson, Richard J},
  journal={arXiv preprint arXiv:2412.10848},
  year={2024}
}

@article{ethos,
  title={Zero shot health trajectory prediction using transformer},
  author={Renc, Pawel and Jia, Yugang and Samir, Anthony E and Was, Jaroslaw and Li, Quanzheng and Bates, David W and Sitek, Arkadiusz},
  journal={NPJ Digital Medicine},
  volume={7},
  number={1},
  pages={256},
  year={2024},
  publisher={Nature Publishing Group UK London}
}

@misc{nhs_sde,
  title = {{Secure Data Environment}},
  author = {{NHS England}},
  year = {2025},
  howpublished = {\url{https://digital.nhs.uk/services/secure-data-environment-service}},
  note = {Accessed: 2025-08-13}
}

@misc{bhf,
  title = {{CVD-COVID-UK / COVID-IMPACT}},
  author = {{British Heart Foundation}},
  howpublished = {\url{https://bhfdatasciencecentre.org/areas/cvd-covid-uk-covid-impact/}},
  year = {2025},
  note = {Accessed: 2025-01-10}
}

@misc{bhf_main,
  title = {{Home- British Heart Foundation}},
  author = {{British Heart Foundation}},
  howpublished = {\url{https://bhfdatasciencecentre.org/}},
  year = {2025},
  note = {Accessed: 2025-07-20}
}

@misc{nhs_digital_agd,
  title = {{Advisory Group for Data (AGD)}},
  author = {{NHS Digital}},
  howpublished = {\url{https://digital.nhs.uk/about-nhs-digital/corporate-information-and-documents/advisory-group-for-data}},
  year = {2025},
  note = {Accessed: 2025-07-20}
}

@misc{nhs_digital_data_access_request,
  title = {{Data Access Request Service (DARS) products and services}},
  author = {{NHS Digital}},
  howpublished = {\url{https://digital.nhs.uk/services/data-access-request-service-dars/dars-products-and-services}},
  year = {2024},
  note = {Accessed: 2025-07-20}
}

@misc{primary_care_data,
  title = {{COVID-19} General Practice Extraction Service (GPES) Data for Pandemic Planning and Research ({GDPPR})},
  author = {{NHS Digital}},
  howpublished = {\url{https://digital.nhs.uk/services/data-access-request-service-dars/dars-products-and-services/data-set-catalogue/gpes-data-for-pandemic-planning-and-research-gdppr}},
  year = {2025},
  note = {Accessed: 2025-08-13}
}

@misc{secondary_care_data,
  title = {{Hospital Episode Statistics (HES)}},
  author = {{NHS Digital}},
  howpublished = {\url{https://digital.nhs.uk/data-and-information/data-tools-and-services/data-services/hospital-episode-statistics}},
  year = {2025},
  note = {Accessed: 2025-08-13}
}

@misc{deaths_data,
  title = {{Civil Registration - Deaths}},
  author = {{Health Data Research Gateway}},
  howpublished = {\url{https://healthdatagateway.org/en/dataset/877}},
  year = {2024},
  note = {Accessed: 2025-08-13}
}

@misc{covid_data_1,
  title = {{COVID-19 Second Generation Surveillance System}},
  author = {{NHS Digital}},
  year = {2023},
  howpublished = {\url{https://digital.nhs.uk/services/data-services-for-commissioners/datasets/covid-19-second-generation-surveillance-system}},
  note = {Accessed: 2025-01-10}
}

@misc{covid_data_2,
  title = {{COVID-19 Vaccination Status}},
  author = {{NHS Digital}},
  howpublished = {\url{https://digital.nhs.uk/services/data-services-for-commissioners/datasets/covid-19-vaccination-status}},
  year = {2023},
  note = {Accessed: 2025-01-10}
}

@article{llama2,
  title={Llama 2: Open foundation and fine-tuned chat models},
  author={Touvron, Hugo and Martin, Louis and Stone, Kevin and Albert, Peter and Almahairi, Amjad and Babaei, Yasmine and Bashlykov, Nikolay and Batra, Soumya and Bhargava, Prajjwal and Bhosale, Shruti and others},
  journal={arXiv preprint arXiv:2307.09288},
  year={2023}
}

@article{phecodes,
  title={Mapping ICD-10 and ICD-10-CM codes to phecodes: workflow development and initial evaluation},
  author={Wu, Patrick and Gifford, Aliya and Meng, Xiangrui and Li, Xue and Campbell, Harry and Varley, Tim and Zhao, Juan and Carroll, Robert and Bastarache, Lisa and Denny, Joshua C and others},
  journal={JMIR medical informatics},
  volume={7},
  number={4},
  pages={e14325},
  year={2019},
  publisher={JMIR Publications Inc., Toronto, Canada}
}

@article{STANDING,
  title={Tackling algorithmic bias and promoting transparency in health datasets: the STANDING Together consensus recommendations},
  author={Alderman, Joseph E and Palmer, Joanne and Laws, Elinor and McCradden, Melissa D and Ordish, Johan and Ghassemi, Marzyeh and Pfohl, Stephen R and Rostamzadeh, Negar and Cole-Lewis, Heather and Glocker, Ben and others},
  journal={The Lancet Digital Health},
  volume={7},
  number={1},
  pages={e64--e88},
  year={2025},
  publisher={Elsevier}
}

@misc{nhs_clinical_classification,
  title = {{Clinical Classifications}},
  author = {{NHS Digital}},
  howpublished = {\url{https://digital.nhs.uk/services/terminology-and-classifications/clinical-classifications}},
  note = {Accessed: 2025-04-14},
  year={2025},
}

@misc{gpdr_data_provision_notice,
  title = {{Data Provision Notice General Practice Data for Planning and Research}},
  author = {{NHS Digital}},
  howpublished = {\url{https://www.spinneybrookmedcentre.co.uk/mf.ashx?ID=36ab53fe-7e31-4383-b4aa-fd525cf8fa50}},
  note = {Accessed: 2026-03-06},
  year={2021},
}

@misc{icd,
  title = {{International Statistical Classification of Diseases and Related Health Problems (ICD)}},
  author = {{World Health Organization}},
  howpublished ={\url{https://www.who.int/standards/classifications/classification-of-diseases}},
  note = {Accessed: 2025-04-14},
  year = {2025},
}

@misc{snomed_ct,
  title = {{SNOMED CT}},
  author = {{NHS Digital}},
  howpublished = {\url{https://digital.nhs.uk/services/terminology-and-classifications/snomed-ct}},
  note = {Accessed: 2025-04-14},
  year = {2025},
}

@misc{g5_cluster,
  title = {{Amazon EC2 G5 Instances}},
  author = {{AWS}},
  howpublished = {\url{https://aws.amazon.com/ec2/instance-types/g5/}},
  year = {2025},
  note = {Accessed: 2025-04-14}
}

@misc{nvidia_a10,
  title = {{NVIDIA A10 Tensor Core GPU}},
  author = {{NVIDIA}},
  howpublished = {\url{https://www.nvidia.com/en-gb/data-center/products/a10-gpu/}},
  year = {2025},
  note = {Accessed: 2025-04-14}
}

@misc{llama2_hf,
  title = {{Llama 2}},
  author = {{Hugging Face}},
  howpublished = {\url{https://huggingface.co/docs/transformers/en/model_doc/llama2}},
  year = {2023},
  note = {Accessed: 2025-04-28}
}

@misc{nanoGPT,
  title = {{nanoGPT}},
  author = {{Andrej Karpathy}},
  howpublished = {\url{https://github.com/karpathy/nanoGPT}},
  year = {2025},
  note = {Accessed: 2025-04-15}
}

@article{rope,
  title={Roformer: Enhanced transformer with rotary position embedding},
  author={Su, Jianlin and Ahmed, Murtadha and Lu, Yu and Pan, Shengfeng and Bo, Wen and Liu, Yunfeng},
  journal={Neurocomputing},
  volume={568},
  pages={127063},
  year={2024},
  publisher={Elsevier}
}

@article{Flashattention_2,
  title={{Flashattention-2}: Faster attention with better parallelism and work partitioning},
  author={Dao, Tri},
  journal={arXiv preprint arXiv:2307.08691},
  year={2023}
}

@article{thygesen2022covid,
  title={{COVID-19} trajectories among 57 million adults in England: a cohort study using electronic health records},
  author={Thygesen, Johan H and Tomlinson, Christopher and Hollings, Sam and Mizani, Mehrdad A and Handy, Alex and Akbari, Ashley and Banerjee, Amitava and Cooper, Jennifer and Lai, Alvina G and Li, Kezhi and others},
  journal={The Lancet Digital Health},
  volume={4},
  number={7},
  pages={e542--e557},
  year={2022},
  publisher={Elsevier}
}

@article{wood2021linked,
  title={Linked electronic health records for research on a nationwide cohort of more than 54 million people in England: data resource},
  author={Wood, Angela and Denholm, Rachel and Hollings, Sam and Cooper, Jennifer and Ip, Samantha and Walker, Venexia and Denaxas, Spiros and Akbari, Ashley and Banerjee, Amitava and Whiteley, William and others},
  journal={BMJ},
  volume={373},
  year={2021},
  publisher={British Medical Journal Publishing Group}
}

@article{pineda2024ethnicity,
  title={Ethnicity data resource in population-wide health records: completeness, coverage and granularity of diversity},
  author={Pineda-Moncus{\'\i}, Marta and Allery, Freya and Delmestri, Antonella and Bolton, Thomas and Nolan, John and Thygesen, Johan H and Handy, Alex and Banerjee, Amitava and Denaxas, Spiros and Tomlinson, Christopher and others},
  journal={Scientific Data},
  volume={11},
  number={1},
  pages={221},
  year={2024},
  publisher={Nature Publishing Group UK London}
}

@article{wilde2023hospital,
  title={Hospital admissions linked to SARS-CoV-2 infection in children and adolescents: cohort study of 3.2 million first ascertained infections in England},
  author={Wilde, Harrison and Tomlinson, Christopher and Mateen, Bilal A and Selby, David and Kanthimathinathan, Hari Krishnan and Ramnarayan, Padmanabhan and Du Pre, Pascale and Johnson, Mae and Pathan, Nazima and Gonzalez-Izquierdo, Arturo and others},
  journal={BMJ},
  volume={382},
  year={2023},
  publisher={British Medical Journal Publishing Group}
}

@article{riley2025uncertainty,
  title={Uncertainty of risk estimates from clinical prediction models: rationale, challenges, and approaches},
  author={Riley, Richard D and Collins, Gary S and Kirton, Laura and Snell, Kym IE and Ensor, Joie and Whittle, Rebecca and Dhiman, Paula and van Smeden, Maarten and Liu, Xiaoxuan and Alderman, Joseph and others},
  journal={BMJ},
  volume={388},
  year={2025},
  publisher={British Medical Journal Publishing Group}
}

@article{probastAI,
  title={PROBAST+ AI: an updated quality, risk of bias, and applicability assessment tool for prediction models using regression or artificial intelligence methods},
  author={Moons, Karel GM and Damen, Johanna AA and Kaul, Tabea and Hooft, Lotty and Navarro, Constanza Andaur and Dhiman, Paula and Beam, Andrew L and Van Calster, Ben and Celi, Leo Anthony and Denaxas, Spiros and others},
  journal={BMJ},
  volume={388},
  year={2025},
  publisher={British Medical Journal Publishing Group}
}

@article{tripodAI,
  title={TRIPOD+ AI statement: updated guidance for reporting clinical prediction models that use regression or machine learning methods},
  author={Collins, Gary S and Moons, Karel GM and Dhiman, Paula and Riley, Richard D and Beam, Andrew L and Van Calster, Ben and Ghassemi, Marzyeh and Liu, Xiaoxuan and Reitsma, Johannes B and Van Smeden, Maarten and others},
  journal={BMJ},
  volume={385},
  year={2024},
  publisher={British Medical Journal Publishing Group}
}

@misc{databricks_14_3_ml,
  title = {{Databricks Runtime 14.3 LTS}},
  author = {{Databricks}},
  howpublished = {\url{https://docs.databricks.com/aws/en/release-notes/runtime/14.3lts}},
  year = {2024},
  note = {Accessed: 2025-04-29}
}

@article{behrt,
  title={{BEHRT}: transformer for electronic health records},
  author={Li, Yikuan and Rao, Shishir and Solares, Jos{\'e} Roberto Ayala and Hassaine, Abdelaali and Ramakrishnan, Rema and Canoy, Dexter and Zhu, Yajie and Rahimi, Kazem and Salimi-Khorshidi, Gholamreza},
  journal={Scientific reports},
  volume={10},
  number={1},
  pages={7155},
  year={2020},
  publisher={Nature Publishing Group UK London}
}

@article{delphi,
  title={Learning the natural history of human disease with generative transformers},
  author={Shmatko, Artem and Jung, Alexander Wolfgang and Gaurav, Kumar and Brunak, S{\o}ren and Mortensen, Laust and Birney, Ewan and Fitzgerald, Tom and Gerstung, Moritz},
  journal={medRxiv},
  year={2024},
  publisher={Cold Spring Harbor Laboratory Press}
}

@article{motor,
  title={MOTOR: a time-to-event foundation model for structured medical records},
  author={Steinberg, Ethan and Fries, Jason and Xu, Yizhe and Shah, Nigam},
  journal={arXiv preprint arXiv:2301.03150},
  year={2023}
}

@article{roc,
  title={{ROC} solid: Receiver operator characteristic ({ROC}) curves as a foundation for better diagnostic tests},
  author={Junge, Mark RJ and Dettori, Joseph R},
  journal={Global spine journal},
  volume={8},
  number={4},
  pages={424--429},
  year={2018},
  publisher={SAGE Publications Sage CA: Los Angeles, CA}
}

@article{average_precision,
  title={Recall, precision and average precision},
  author={Zhu, Mu},
  journal={Department of Statistics and Actuarial Science, University of Waterloo, Waterloo},
  volume={2},
  number={30},
  pages={6},
  year={2004}
}

@article{brier_score,
  title={Verification of forecasts expressed in terms of probability},
  author={Brier, Glenn W. and others},
  journal={Monthly weather review},
  volume={78},
  number={1},
  pages={1--3},
  year={1950},
  publisher={War Department, Office of the Chief Signal Officer}
}

@misc{log_regression,
  title = {{LogisticRegression}},
  author = {{scikit-learn}},
  howpublished = {\url{https://scikit-learn.org/stable/modules/generated/sklearn.linear_model.LogisticRegression.html}},
  year = {2025},
  note = {Accessed: 2025-08-13}
}

@misc{xg_boost,
  title = {{XGBoost Documentation}},
  author = {{DMLC XGBoost}},
  howpublished = {\url{https://xgboost.readthedocs.io/en/stable/index.html}},
  year = {2022},
  note = {Accessed: 2025-07-14}
}

@article{qrisk,
  title={Development and validation of {QRISK3} risk prediction algorithms to estimate future risk of cardiovascular disease: prospective cohort study},
  author={Hippisley-Cox, Julia and Coupland, Carol and Brindle, Peter},
  journal={BMJ},
  volume={357},
  year={2017},
  publisher={British Medical Journal Publishing Group}
}

@article{chad_vasc,
  title={Refining clinical risk stratification for predicting stroke and thromboembolism in atrial fibrillation using a novel risk factor-based approach: the euro heart survey on atrial fibrillation},
  author={Lip, Gregory YH and Nieuwlaat, Robby and Pisters, Ron and Lane, Deirdre A and Crijns, Harry JGM},
  journal={Chest},
  volume={137},
  number={2},
  pages={263--272},
  year={2010},
  publisher={Elsevier}
}

@misc{llama_4,
  title = {{The Llama 4 herd: The beginning of a new era of natively multimodal AI innovation}},
  author = {{Meta}},
  howpublished = {\url{https://ai.meta.com/blog/llama-4-multimodal-intelligence/}},
  year = {2025},
  note = {Accessed: 2025-07-14}
}

@article{medbert,
  title={Med-BERT: pretrained contextualized embeddings on large-scale structured electronic health records for disease prediction},
  author={Rasmy, Laila and Xiang, Yang and Xie, Ziqian and Tao, Cui and Zhi, Degui},
  journal={NPJ digital medicine},
  volume={4},
  number={1},
  pages={86},
  year={2021},
  publisher= {Nature Publishing Group UK London}
}

@misc{copi,
  title = {{Control of patient information (COPI) notice}},
  author = {{NHS Digital}},
  howpublished = {\url{https://digital.nhs.uk/coronavirus/coronavirus-covid-19-response-information-governance-hub/control-of-patient-information-copi-notice}},
  note = {Accessed: 2025-07-14},
  year = {2022}
}

@article{ehr_scaling,
  title={Exploring Scaling Laws for {EHR} Foundation Models},
  author={Zhang, Sheng and Liu, Qin and Usuyama, Naoto and Wong, Cliff and Naumann, Tristan and Poon, Hoifung},
  journal={arXiv preprint arXiv:2505.22964},
  year={2025}
}

@article{context_clues,
  title={Context clues: Evaluating long context models for clinical prediction tasks on ehrs},
  author={Wornow, Michael and Bedi, Suhana and Hernandez, Miguel Angel Fuentes and Steinberg, Ethan and Fries, Jason Alan and R{\'e}, Christopher and Koyejo, Sanmi and Shah, Nigam H},
  journal={arXiv preprint arXiv:2412.16178},
  year={2024}
}

@article{mimic_iv,
  title={MIMIC-IV, a freely accessible electronic health record dataset},
  author={Johnson, Alistair EW and Bulgarelli, Lucas and Shen, Lu and Gayles, Alvin and Shammout, Ayad and Horng, Steven and Pollard, Tom J and Hao, Sicheng and Moody, Benjamin and Gow, Brian and others},
  journal={Scientific data},
  volume={10},
  number={1},
  pages={1},
  year={2023},
  publisher={Nature Publishing Group UK London}
}

@article{document_masking,
  title={Analysing the impact of sequence composition on language model pre-training},
  author={Zhao, Yu and Qu, Yuanbin and Staniszewski, Konrad and Tworkowski, Szymon and Liu, Wei and Mi{\l}o{\'s}, Piotr and Wu, Yuxiang and Minervini, Pasquale},
  journal={arXiv preprint arXiv:2402.13991},
  year={2024}
}

@misc{bmj_foresight,
  title={NHS England faces investigation over granting Foresight access to GP patient data},
  author={Armstrong, Stephen},
  year={2025},
  publisher={British Medical Journal Publishing Group}
}

@article{indirect_covid_1,
  title={Estimating excess 1-year mortality associated with the {COVID-19} pandemic according to underlying conditions and age: a population-based cohort study},
  author={Banerjee, Amitava and Pasea, Laura and Harris, Steve and Gonzalez-Izquierdo, Arturo and Torralbo, Ana and Shallcross, Laura and Noursadeghi, Mahdad and Pillay, Deenan and Sebire, Neil and Holmes, Chris and others},
  journal={The lancet},
  volume={395},
  number={10238},
  pages={1715--1725},
  year={2020},
  publisher={Elsevier}
}

@article{indirect_covid_2,
  title={Monitoring indirect impact of {COVID-19} pandemic on services for cardiovascular diseases in the UK},
  author={Ball, Simon and Banerjee, Amitava and Berry, Colin and Boyle, Jonathan R and Bray, Benjamin and Bradlow, William and Chaudhry, Afzal and Crawley, Rikki and Danesh, John and Denniston, Alastair and others},
  journal={Heart},
  volume={106},
  number={24},
  pages={1890--1897},
  year={2020},
  publisher={BMJ Publishing Group Ltd and British Cardiovascular Society}
}

@article{fpr10,
  title={Proteomic signatures improve risk prediction for common and rare diseases},
  author={Carrasco-Zanini, Julia and Pietzner, Maik and Davitte, Jonathan and Surendran, Praveen and Croteau-Chonka, Damien C and Robins, Chloe and Torralbo, Ana and Tomlinson, Christopher and Gr{\"u}nschl{\"a}ger, Florian and Fitzpatrick, Natalie and others},
  journal={Nature medicine},
  volume={30},
  number={9},
  pages={2489--2498},
  year={2024},
  publisher={Nature Publishing Group US New York}
}

@article{handy2021,
  title={A nationwide deep learning pipeline to predict stroke and {COVID-19} death in atrial fibrillation},
  author={Handy, Alex and Wood, Angela and Sudlow, Cathie and Tomlinson, Christopher and Kee, Frank and Thygesen, Johan H and Mamouei, Mohammad and Sofat, Reecha and Dobson, Richard and Ip, Samantha and others},
  journal={Medrxiv},
  year={2021},
  publisher={Cold Spring Harbor Laboratory Press}
}

@article{pr_curve_w_imbalance_1,
  title={The precision--recall curve overcame the optimism of the receiver operating characteristic curve in rare diseases},
  author={Ozenne, Brice and Subtil, Fabien and Maucort-Boulch, Delphine},
  journal={Journal of clinical epidemiology},
  volume={68},
  number={8},
  pages={855--859},
  year={2015},
  publisher={Elsevier}
}

@article{pr_curve_w_imbalance_2,
  title={The precision-recall plot is more informative than the {ROC} plot when evaluating binary classifiers on imbalanced datasets},
  author={Saito, Takaya and Rehmsmeier, Marc},
  journal={PloS one},
  volume={10},
  number={3},
  pages={e0118432},
  year={2015},
  publisher={Public Library of Science San Francisco, CA USA}
}

@article{allery2023,
  title={Towards mitigating health inequity via machine learning: a nationwide cohort study to develop and validate ethnicity-specific models for prediction of cardiovascular disease risk in {COVID-19} patients},
  author={Allery, Freya and Pineda-Moncus{\'\i}, Marta and Tomlinson, Christopher and Pontikos, Nikolas and Thygesen, Johan H and Khalid, Sara and {{CVD-COVID-UK/COVID-IMPACT Consortium}}},
  journal={medRxiv},
  year={2023},
  publisher={Cold Spring Harbor Laboratory Press}
}

@article{thygesen2025rare,
  title={Prevalence and demographics of 331 rare diseases and associated {COVID-19}-related mortality among 58 million individuals: a nationwide retrospective observational study},
  author={Thygesen, Johan H and Zhang, Huayu and Issa, Hanane and Wu, Jinge and Hama, Tuankasfee and Phiho-Gomes, Ana-Caterina and Groza, Tudor and Khalid, Sara and Lumbers, Thomas R and Hocaoglu, Mevhibe and others},
  journal={The Lancet Digital Health},
  volume={7},
  number={2},
  pages={e145--e156},
  year={2025},
  publisher={Elsevier}
}

@article{arora2023,
  title={The value of standards for health datasets in artificial intelligence-based applications},
  author={Arora, Anmol and Alderman, Joseph E and Palmer, Joanne and Ganapathi, Shaswath and Laws, Elinor and Mccradden, Melissa D and Oakden-Rayner, Lauren and Pfohl, Stephen R and Ghassemi, Marzyeh and Mckay, Francis and others},
  journal={Nature medicine},
  volume={29},
  number={11},
  pages={2929--2938},
  year={2023},
  publisher={Nature Publishing Group US New York}
}

@inproceedings{clip,
  title={Learning transferable visual models from natural language supervision},
  author={Radford, Alec and Kim, Jong Wook and Hallacy, Chris and Ramesh, Aditya and Goh, Gabriel and Agarwal, Sandhini and Sastry, Girish and Askell, Amanda and Mishkin, Pamela and Clark, Jack and others},
  booktitle={International conference on machine learning},
  pages={8748--8763},
  year={2021},
  organization={PmLR}
}

@inproceedings{perceiver,
  title={Perceiver: General perception with iterative attention},
  author={Jaegle, Andrew and Gimeno, Felix and Brock, Andy and Vinyals, Oriol and Zisserman, Andrew and Carreira, Joao},
  booktitle={International conference on machine learning},
  pages={4651--4664},
  year={2021},
  organization={PMLR}
}

@article{pineda2025disparity,
  title={Ethnic disparities in {COVID-19} mortality and cardiovascular disease in {England} and {Wales} between 2020--2022},
  author={Pineda-Moncus{\'\i}, Marta and Allery, Freya and Abbasizanjani, Hoda and Powell, David and Prats-Uribe, Albert and Thygesen, Johan H and Wood, Angela and Tomlinson, Christopher and Banerjee, Amitava and Akbari, Ashley and others},
  journal={Nature Communications},
  volume={16},
  number={1},
  pages={6059},
  year={2025},
  publisher={Nature Publishing Group UK London}
}

@article{sjoding2020spo2,
  title={Racial bias in pulse oximetry measurement},
  author={Sjoding, Michael W and Dickson, Robert P and Iwashyna, Theodore J and Gay, Steven E and Valley, Thomas S},
  journal={New England Journal of Medicine},
  volume={383},
  number={25},
  pages={2477--2478},
  year={2020},
  publisher={Mass Medical Soc}
}

@article{kostopoulou2021,
  title={Can decision support combat incompleteness and bias in routine primary care data?},
  author={Kostopoulou, Olga and Tracey, Christopher and Delaney, Brendan C},
  journal={Journal of the American Medical Informatics Association},
  volume={28},
  number={7},
  pages={1461--1467},
  year={2021},
  publisher={Oxford Academic}
}

@article{zghebi2022clinical,
  title={Clinical code usage in UK general practice: a cohort study exploring 18 conditions over 14 years},
  author={Zghebi, Salwa S and Reeves, David and Grigoroglou, Christos and McMillan, Brian and Ashcroft, Darren M and Parisi, Rosa and Kontopantelis, Evangelos},
  journal={BMJ Open},
  volume={12},
  number={7},
  pages={e051456},
  year={2022},
  publisher={British Medical Journal Publishing Group}
}

@article{lancet_rheum,
  title   = {Translating AI innovation into clinical practice},
  author  = {{The Lancet Rheumatology}},
  journal = {The Lancet Rheumatology},
  year    = {2025},
  volume  = {7},
  number  = {7},
  pages   = {e451},
  doi     = {10.1016/S2665-9913(25)00161-4},
  url     = {https://doi.org/10.1016/S2665-9913(25)00161-4},
  publisher = {Elsevier},
  issn    = {2665-9913},
  note    = {Published July 1, 2025}
}

@article{medtok,
  title={Multimodal medical code tokenizer},
  author={Su, Xiaorui and Messica, Shvat and Huang, Yepeng and Johnson, Ruth and Fesser, Lukas and Gao, Shanghua and Sahneh, Faryad and Zitnik, Marinka},
  journal={arXiv preprint arXiv:2502.04397},
  year={2025}
}

@article{epic_ehr_model,
  title={Generative medical event models improve with scale},
  author={Waxler, Shane and Blazek, Paul and White, Davis and Sneider, Daniel and Chung, Kevin and Nagarathnam, Mani and Williams, Patrick and Voeller, Hank and Wong, Karen and Swanhorst, Matthew and others},
  journal={arXiv preprint arXiv:2508.12104},
  year={2025}
}

@article{knight2020risk,
  title={Risk stratification of patients admitted to hospital with covid-19 using the ISARIC WHO Clinical Characterisation Protocol: development and validation of the 4C Mortality Score},
  author={Knight, Stephen R and Ho, Antonia and Pius, Riinu and Buchan, Iain and Carson, Gail and Drake, Thomas M and Dunning, Jake and Fairfield, Cameron J and Gamble, Carrol and Green, Christopher A and others},
  journal={BMJ},
  volume={370},
  year={2020},
  publisher={British Medical Journal Publishing Group}
}

@article{logan2025early,
  title={The early warning paradox},
  author={Logan Ellis, Hugh and Palmer, Edward and Teo, James T and Whyte, Martin and Rockwood, Kenneth and Ibrahim, Zina},
  journal={npj Digital Medicine},
  volume={8},
  number={1},
  pages={81},
  year={2025},
  publisher={Nature Publishing Group UK London}
}

@inproceedings{carlini2021extracting,
  title={Extracting training data from large language models},
  author={Carlini, Nicholas and Tramer, Florian and Wallace, Eric and Jagielski, Matthew and Herbert-Voss, Ariel and Lee, Katherine and Roberts, Adam and Brown, Tom and Song, Dawn and Erlingsson, Ulfar and others},
  booktitle={30th USENIX security symposium (USENIX Security 21)},
  pages={2633--2650},
  year={2021}
}

@article{grainmatter,
  title={GRAIMATTER green paper: Recommendations for disclosure control of trained machine learning (ML) models from trusted research environments (TREs)},
  author={Jefferson, Emily and Liley, James and Malone, Maeve and Reel, Smarti and Crespi-Boixader, Alba and Kerasidou, Xaroula and Tava, Francesco and McCarthy, Andrew and Preen, Richard and Blanco-Justicia, Alberto and others},
  journal={arXiv preprint arXiv:2211.01656},
  year={2022}
}

@article{llms_are_ehr_encoders,
  title={Large language models are powerful electronic health record encoders},
  author={Hegselmann, Stefan and von Arnim, Georg and Rheude, Tillmann and Kronenberg, Noel and Sontag, David and Hindricks, Gerhard and Eils, Roland and Wild, Benjamin},
  journal={arXiv preprint arXiv:2502.17403},
  year={2025}
}

@article{knight2022association,
  title={Association of {COVID-19} with major arterial and venous thrombotic diseases: a population-wide cohort study of 48 million adults in England and Wales},
  author={Knight, Rochelle and Walker, Venexia and Ip, Samantha and Cooper, Jennifer A and Bolton, Thomas and Keene, Spencer and Denholm, Rachel and Akbari, Ashley and Abbasizanjani, Hoda and Torabi, Fatemeh and others},
  journal={Circulation},
  volume={146},
  number={12},
  pages={892--906},
  year={2022},
  publisher={Lippincott Williams \& Wilkins Hagerstown, MD}
}

@misc{indirect_covid_3,
  title = {{Managing NHS backlogs and waiting times in England}},
  author = {{National Audit Office}},
  year = {2022},
  howpublished = {\url{https://www.nao.org.uk/reports/managing-nhs-backlogs-and-waiting-times-in-england/}},
  note = {Accessed: 2026-05-25}
}

@misc{coronavirus_infection_survey,
  title = {{Coronavirus ({COVID-19}) Infection Survey, antibody data, UK: 18 May 2022}},
  author = {{Office for National Statistics}},
  year = {2022},
  howpublished ={\url{https://www.ons.gov.uk/peoplepopulationandcommunity/healthandsocialcare/conditionsanddiseases/bulletins/coronaviruscovid19infectionsurveyantibodyandvaccinationdatafortheuk/18may2022}},
  note = {Accessed: 2026-05-25}
}

@article{lillie2020novel,
  title={Novel coronavirus disease ({Covid-19}): The first two patients in the {UK} with person to person transmission},
  author={Lillie, Patrick J and Samson, Alison and Li, Angharad and Adams, Kate and Capstick, Richard and Barlow, Gavin D and Easom, Nicholas and Hamilton, Eve and Moss, Peter J and Gow, Ashley and others},
  journal={Journal of Infection},
  volume={80},
  number={5},
  pages={578--579},
  year={2020},
  publisher={Elsevier},
  doi={10.1016/j.jinf.2020.02.020}
}

@techreport{cabinetoffice2022living,
  title={{COVID-19} Response: Living with {COVID-19}},
  author={{Cabinet Office}},
  institution={{UK Government}},
  year={2022},
  address={London},
  url={https://www.gov.uk/government/publications/covid-19-response-living-with-covid-19},
  note={Accessed: 2026-05-26} 
}

\newpage
\appendix
\section{Training}
\label{section_methods_training_append}

\subsection{Objective}
\label{section_training_obj}
Foresight-E was trained with an autoregressive next-token prediction objective, analogous to the paradigm used for LLMs \cite{gpt3,foresight,ethos}. Let $N$ be the batch size, $T$ the maximum sequence length, and $V$ the vocabulary size. At each position $t$, given preceding tokens $y_{n,<t}$, the model outputs a distribution over $V$. A binary mask $m_{n,t}$ excludes padding and the initial demographic tokens from loss calculation. Excluding the initial tokens aims to mitigate the risk of biasing the model by learning to predict based solely on demographic features. The loss, $\mathcal{L}$, given the model weights, $\theta$ and the true token, $y_{n,t}$, is:

\begin{equation}
\mathcal{L} = -\frac{1}{\sum_{n,t} m_{n,t}} 
\sum_{n=1}^{N} \sum_{t=1}^{T} m_{n,t} \log P(y_{n,t} \mid y_{n,<t}, \theta)
\end{equation}

\subsection{Architecture} 
We adapted the Llama 2 transformer-decoder \cite{llama2,llama2_hf} architecture with Rotary Positional Embeddings (ROPE) \cite{rope} and FlashAttention-2 \cite{Flashattention_2}. Pretrained weights were not imported into the SDE due to governance restrictions and the custom vocabulary; therefore, training was conducted from scratch. We scaled the Llama decoder architecture down to 243 million parameters to facilitate efficient training on NVIDIA A10 GPUs. While we maintained the foundational hyperparameter ratios used across the Llama family \cite{llama2}, we adapted the final configuration to a 12-layer architecture with a hidden size of 1024, 8 attention heads, a feed-forward dimension of 4096, and a 1024-token context window. Additionally, input and output embeddings were tied to optimise parameter efficiency.

\subsection{Training Protocol} 
We used bfloat16 mixed precision, attention dropout 0.1, gradient clipping (max norm 1.0), and weight decay 0.1. The Adam optimizer had $\beta_1 = 0.9$, $\beta_2 = 0.95$, with a linear warm-up over 3\% of steps to a peak learning rate of $5\times 10^{-4}$, then cosine decay. The global batch size was 128, achieved via 8-way data parallelism and gradient accumulation (factor 2). Sequences were right-padded to 1024 tokens. The model was trained for one epoch, with validation loss evaluated every 1{,}000 steps on 32k sampled sequences. Training was completed in 4 days of wall-clock compute time distributed across 8 NVIDIA A10 GPUs. We did not conduct hyperparameter tuning due to resource constraints; instead, training parameters were selected based on standard defaults from prior literature \cite{ foresight, nanoGPT}. 

\newpage
\section{Endpoint to Token Mapping}

\begin{tabularx}{\textwidth}{cX}
\caption{Mapping of evaluation endpoints and associated figures to timeline tokens. An example phecode definition is shown.}
\label{table_eval_endpoints}\\
\toprule
\textbf{Endpoint} & \textbf{Tokens} \\
\midrule
Mortality & \texttt{DEATH} \\
Hospitalisation & \texttt{APC} \\
Emergency Hospitalisation & \texttt{APC\_ADMIMETH\_21}, \texttt{APC\_ADMIMETH\_22}, \texttt{APC\_ADMIMETH\_23}, \texttt{APC\_ADMIMETH\_24}, \texttt{APC\_ADMIMETH\_25}, \texttt{APC\_ADMIMETH\_28}, \texttt{APC\_ADMIMETH\_2A}, \texttt{APC\_ADMIMETH\_2B}, \texttt{APC\_ADMIMETH\_2C}, \texttt{APC\_ADMIMETH\_2D}  \\
Phecode 8.0 - Intestinal Infection & 
\texttt{ICD10\_A000}, \texttt{ICD10\_A009}, \texttt{ICD10\_A011}, \texttt{ICD10\_A012}, \texttt{ICD10\_A013}, \texttt{ICD10\_A014}, 
\texttt{ICD10\_A059}, \texttt{ICD10\_A060}, \texttt{ICD10\_A062}, \texttt{ICD10\_A063}, \texttt{ICD10\_A064}, \texttt{ICD10\_A065}, \texttt{ICD10\_A067}, \texttt{ICD10\_A068}, \texttt{ICD10\_A069}, \texttt{ICD10\_A079} \\
\bottomrule
\end{tabularx}

\newpage
\section{TRIPOD+AI Checklist}
\label{section_tripod}

\begin{tabularx}{\textwidth}{p{2cm}Xc}
\caption{Assessment of Foresight-E against the Transparent Reporting of a multivariable prediction model for Individual Prognosis Or Diagnosis (TRIPOD)+AI Checklist \cite{tripodAI}. The results section is excluded as not included in this paper.}
\label{table_tripod}\\
\toprule
\textbf{Item} & \textbf{Checklist item} & \textbf{Section}\\
\midrule

\multicolumn{3}{l}{\textbf{Title}} \\
\midrule
Title & Identify the study as developing or evaluating the performance of a multivariable prediction model, the target population, and the outcome to be predicted & Title\\
\midrule
\multicolumn{3}{l}{\textbf{Abstract}} \\
\midrule
Title & Identify the study as developing or evaluating the performance of a multivariable prediction model, the target population, and the outcome to be predicted & Title\\
Background&Provide a brief explanation of the healthcare context and rationale for developing or evaluating the performance of all models & Abstract\\
Objectives&Specify the study objectives, including whether the study describes model development, evaluation, or both & Abstract\\
Methods&Describe the sources of data & Abstract\\
& Describe the eligibility criteria and setting where the data were collected& Abstract\\
& Specify the outcome to be predicted by the model, including time horizon of predictions in case of prognostic models& Abstract\\
& Specify the type of model, a summary of the model-building steps, and the method for internal validation& Abstract\\
& Specify the measures used to assess model performance (eg, discrimination, calibration, clinical utility)& Abstract\\
Results& Report the number of participants and outcome events& Pending\\
& Summarise the predictors in the final model& Pending\\
& Report model performance estimates (with confidence intervals)& Pending\\
Discussion & Give an overall interpretation of the main results& Pending\\
Registration & Give the registration number and name of the registry or repository& None\\

\midrule
\multicolumn{3}{l}{\textbf{Introduction}} \\
\midrule
Background & Explain the healthcare context (including whether diagnostic or prognostic) and rationale for developing or evaluating the prediction model, including references to existing models & \ref{section_intro}\\
 & Describe the target population and the intended purpose of the prediction model in the context of the care pathway, including its intended users (eg, healthcare professionals, patients, public)
 & \ref{section_intro}\\
 & Describe any known health inequalities between sociodemographic groups & \ref{section_intro}\\
Objectives & Specify the study objectives, including whether the study describes the development or validation of a prediction model (or both) & \ref{section_intro}\\

\midrule
\multicolumn{3}{l}{\textbf{Methods}} \\
\midrule
Data & Describe the sources of data separately for the development and evaluation datasets (eg, randomised trial, cohort, routine care or registry data), the rationale for using these data, and representativeness of the data & 
\ref{section_methods_data} \\
 & Specify the dates of the collected participant data, including start and end of participant accrual; and, if applicable, end of follow-up
 & \ref{section_methods_data}\\
Participants & Specify key elements of the study setting (eg, primary care, secondary care, general population) including the number and location of centres
 & \ref{section_methods_data} \\
 & Describe the eligibility criteria for study participants & \ref{section_methods_data}\\
 & Give details of any treatments received, and how they were handled during model development or evaluation, if relevant & \ref{section_methods_timelines} \\
Data
preparation & Describe any data pre-processing and quality checking, including whether this was similar across relevant sociodemographic groups & \ref{section_methods_data}, \ref{section_methods_timelines}\\
Outcome & Clearly define the outcome that is being predicted and the time horizon, including how and when assessed, the rationale for choosing this outcome, and whether the method of outcome assessment is consistent across sociodemographic groups
 & \ref{sect:method_outcomes}\\
 & If outcome assessment requires subjective interpretation, describe the qualifications and demographic characteristics of the outcome assessors
 & N/A \\
 & Report any actions to blind assessment of the outcome to be predicted & None\\
Predictors & Describe the choice of initial predictors (eg, literature, previous models, all available predictors) and any pre-selection of predictors before model building
 & \ref{section_methods_data}, \ref{section_methods_timelines}\\
 & Clearly define all predictors, including how and when they were measured (and any actions to blind assessment of predictors for the outcome and other predictors) & \ref{section_methods_data}, \ref{section_tokenisation}\\
 & If predictor measurement requires subjective interpretation, describe the qualifications and demographic characteristics of the predictor assessors & N/A\\
Sample size & Explain how the study size was arrived at (separately for development and evaluation), and justify that the study size was sufficient to answer the research question. Include details of any sample size calculation & \ref{section_methods_data}\\
Missing data & Describe how missing data were handled. Provide reasons for omitting any data & \ref{section_methods_data}, \ref{section_tokenisation}\\
Analytical methods & Describe how the data were used (eg, for development and evaluation of model performance) in the analysis, including whether the data were partitioned, considering any sample size requirements & \ref{section_methods_data}\\
 & Depending on the type of model, describe how predictors were handled in the analyses (functional form, rescaling, transformation, or any standardisation) & \ref{section_methods_data}, \ref{section_methods_timelines}, \ref{section_tokenisation} \\
 & Specify the type of model, rationale†, all model-building steps, including any hyperparameter tuning, and method for internal validation
 & \ref{section_methods_training} \\
 & Describe if and how any heterogeneity in estimates of model parameter values and model performance was handled and quantified across clusters (eg, hospitals, countries). See TRIPOD-Cluster for additional considerations & None\\
 & Specify all measures and plots used (and their rationale) to evaluate model performance (eg, discrimination, calibration,
clinical utility) and, if relevant, to compare multiple models & \ref{sect:method_eval_metrics}, \ref{section_baseline}\\
 & Specify all measures and plots used (and their rationale) to evaluate model performance (eg, discrimination, calibration, clinical utility) and, if relevant, to compare multiple models & \ref{sect:method_eval_metrics} \\
 & Describe any model updating (eg, recalibration) arising from the model evaluation, either overall or for particular socio-demographic groups or settings & None \\
 & For model evaluation, describe how the model predictions were calculated (eg, formula, code, object, application programming interface)
 & \ref{section_inference}\\
Class imbalance & If class imbalance methods were used, state why and how this was done, and any subsequent methods to recalibrate the model or the model predictions & None \\
Fairness & Describe any approaches that were used to address model fairness and their rationale & \ref{section_methods_data}, \ref{section_training_obj}\\
Model output & Specify the output of the prediction model (eg, probabilities, classification). Provide details and rationale for any classification and how the thresholds were identified & \ref{section_inference}, \ref{sect:method_eval_metrics}\\
Training versus evaluation & Identify any differences between the development and evaluation data in healthcare setting, eligibility criteria, outcome, and predictors & \ref{section_methods_data}, \ref{section_tokenisation},\ref{sect:method_eval_metrics}\\
Ethical approval & Name the institutional research board or ethics committee that approved the study and describe the participant informed consent or the ethics committee waiver of informed consent
 & \ref{subsection_goverance}\\

\midrule
\multicolumn{3}{l}{\textbf{Open Science}} \\
\midrule
Funding & Give the source of funding and the role of the funders for the present study & Blinded for review \\
Conflicts of
interest & Declare any conflicts of interest and financial disclosures for all authors & Blinded for review \\
Protocol & Indicate where the study protocol can be accessed or state that a protocol was not prepared & Currently not publicly released \\
Registration & Provide registration information for the study, including register name and registration number, or state that the study was not registered & Not registered\\
Data sharing & Provide details of the availability of the study data & \ref{section_data_availability}\\
Code sharing & Provide details of the availability of the analytical code & \ref{section_code_availability}\\
\midrule
\multicolumn{3}{l}{\textbf{Patient and public involvement}} \\
\midrule
Patient
and public
involvement & Provide details of any patient and public involvement during the design, conduct, reporting, interpretation, or dissemination of the study or state no involvement & \ref{subsection_goverance}\\

\midrule
\multicolumn{3}{l}{\textbf{Discussion}} \\
\midrule
Interpretation & Give an overall interpretation of the main results, including issues of fairness in the context of the objectives and previous studies
 & Pending \\
Limitations & Discuss any limitations of the study (such as a non-representative sample, sample size, overfitting, missing data) and their effects on any biases, statistical uncertainty, and generalisability
 & \ref{sect:dis_data}\\
Usability of the model in the context of current care & Describe how poor quality or unavailable input data (eg, predictor values) should be assessed and handled when implementing the prediction model & N/A \\
 & Specify whether users will be required to interact in the handling of the input data or use of the model, and what level of expertise is required of users
 & \ref{sect:diss_use_cases}\\
 & Discuss any next steps for future research, with a specific view to applicability and generalisability of the model & \ref{sect:diss}\\

\bottomrule
\end{tabularx}

\pagebreak
\section{PROBAST+AI Assessment}
\label{section_PROBAST}

\begin{tabularx}{\textwidth}{p{3cm}Xl}
\caption{Assessment of Foresight-E against Risk Of Bias ASsessment Tool (PROBAST) + AI tool \cite{probastAI}}
\label{table_problast}\\
\toprule
\textbf{Item} & \textbf{Description} &
\makecell{\textbf{Section/} \\ \textbf{Comment}}\\

\midrule
\multicolumn{3}{l}{\textbf{Step 1: PICOTS guidance}} \\
\midrule
Population & Define the target population (e.g., patients) in whom the assessed prediction models are to be applied. The target population not only directs search strings and in/exclusion criteria of prediction models or prediction model studies in case of a systematic literature review, but also directs the applicability assessment. & \ref{section_methods_data} \\
Index Model & Define the targeted prediction models to be assessed, which may be a single prediction model (the index model) of which the predictive accuracy is meta-analysed across multiple external evaluation studies of that index model but may also address multiple prediction models (developed or evaluated) for the targeted population, outcome or setting, depending on the assessor’s or prediction model review focus. & \ref{section_methods_training} \\
Comparator model(s) & Define the other prediction models whose predictive ability is compared to that of the index model. & \ref{section_baseline} \\
Outcome(s) & Define the outcomes or endpoints that are predicted by the index (and possibly comparator) prediction models in the target population. & \ref{sect:method_outcomes}\\
Timing & Define the moment or time-point (e.g., in the patient work-up) at which the prediction with the prediction models is made (i.e., the start point or T0 of the use of the models). & \ref{sect:method_outcomes} \\
& Define the time or follow-up period in which the outcomes are being predicted by the prediction models in the targeted population (prediction horizon). & \ref{sect:method_outcomes} \\
Setting and intended use
of the prediction model
& Define the healthcare setting or context to which the index prediction models apply. The prediction ability of models may change across healthcare settings or contexts. & \ref{section_intro} \\

\midrule
\multicolumn{3}{l}{\textbf{Step 2: Classify the type of prediction model assessment}} \\
\midrule

Development only & Prediction model development only, i.e., without evaluation of its performance. & \checkmark\\
Evaluation only & External validation of one or more existing models in new data \\
Combination & Prediction model development combined in the same study(publication) with the evaluation of its apparent performance, internal validation performance, or external validation performance. \\

\midrule
\multicolumn{3}{l}{\textbf{Step 3: Assess quality and applicability or risk of bias and applicability}} \\
\midrule
Participants and data sources & Describe the sources of data and criteria for participant selection &  \ref{section_methods_data} \\
&Were appropriate data sources used? & Yes \\
&Was an appropriate study design used? & Yes\\
&Did the in- and exclusions of study participants result in a representative
dataset? & Yes\\
& Concern regarding quality of selection of participants and data sources &  Low \\
& Concern that the (data of the) included participants do not match the review
question or the assessor’s intended use of the prediction model &  Low \\

Predictors & List and describe predictors included in the final prediction model, how they were defined and assessed, and their
timing of assessment & \ref{section_methods_data}, \ref{section_tokenisation} \\
& Were predictors defined and assessed in a similar way for all participants? & Yes \\
& Was any pre-processing of predictors similar for all participants? & Yes \\
& Were predictor assessments made without knowledge of outcome data? & Yes \\
& Were the predictors included in the model available at the time the model was intended to be used? & Yes \\
& Concern regarding the quality of the predictors or their assessment & Low \\
& Concern that the definition, pre-processing, assessment, or timing of
assessment of the predictors in the model do not match the review question or
the assessor’s intended use & Low \\

Outcome & Describe the outcome, how it was defined and determined, and the time interval between predictor assessment
and outcome determination & \ref{sect:method_outcomes} \\
& At what time point was the outcome determined?
If a composite outcome was used, describe the relative frequency/distribution of each contributing outcome? & \ref{sect:method_outcomes}  \\
& Were outcomes defined and assessed appropriately? & Yes \\
& Were outcomes defined and assessed in a similar way for all participants? & Yes \\ 
& Were outcome assessments made without use or knowledge of predictor
data? & Yes \\
& Was the time interval between predictor assessment and outcome
assessment appropriate? & Yes \\
& Concern regarding quality of the outcome or its determination & Low \\
& Concern that the outcome, its definition, assessment, or timing of assessment
do not match the review question or the assessor’s intended use & Low \\

Analysis & Describe the numbers of participants, number of candidate predictors, number of outcome events & \ref{section_methods_data}, \ref{sect:method_outcomes}\\
& Describe how the prediction model was developed (e.g., with respect to modelling technique,
predictor selection, and classification or risk group definition) & \ref{section_methods_training}\\
& Describe the performance measures of the prediction model, e.g., (re)calibration, discrimination,
(re)classification, net benefit, and whether they were adjusted for optimism & \ref{sect:method_eval_metrics} \\
& Describe missing data on predictors and outcomes as well as methods used for handling
these missing data & \ref{section_tokenisation} \\
& Was there evidence that the sample size was reasonable? & Yes \\
& Were continuous and categorical predictors handled appropriately? & Yes \\
& Were participants with missing or censored data handled appropriately in
the analysis? & N/A \\
& If methods to address class imbalance were used, was the model or the model
predictions recalibrated? & N/A \\
& Were methods used to address potential model overfitting? & Yes \\
& Concern regarding quality of the analysis & Low \\

\midrule
\multicolumn{3}{l}{\textbf{Step 4: Assess the overall concerns regarding quality, risk of bias and applicability of the prediction model}} \\
\midrule

Overall concern regarding quality of the prediction model development & Low concern regarding quality- If all four domains were rated low concern regarding quality. & \checkmark \\
& High concern regarding quality- If at least one domain was rated high concern regarding quality. & \\
& Unclear concern regarding quality- If at least one domain was rated unclear concern regarding quality and no domains were rated high concern. \\

Overall concern regarding applicability of the prediction model development & Low concern for applicability- If all three domains were rated low concern for applicability. & \checkmark \\
& High concern for applicability- If at least one domain was rated high concern for applicability. & \\
& Unclear concern for applicability- If at least one domain was rated unclear concern for applicability and no domains were rated high concern. & \\
\bottomrule
\end{tabularx}

\end{document}